%% file: main.tex
\documentclass[DIV15, a4paper]{scrartcl}

\usepackage[ruled, linesnumbered, vlined, commentsnumbered]{algorithm2e}
\usepackage{authblk}
\usepackage{prettyref}
\usepackage{graphicx}
\usepackage{subfig}
\usepackage{booktabs}
\usepackage{colortbl}
\usepackage{amssymb}
\usepackage{amsfonts}
\usepackage{amsmath}
\usepackage{amsthm}
\usepackage{multirow}
\usepackage{tabularx}
\usepackage{footnote}
\usepackage{threeparttable}
\usepackage{amsopn}
\usepackage{hhline}
\usepackage{cite}
\usepackage{subfig}
\usepackage{setspace}

\usepackage{tikz}
\usepackage{pgfplots}
\usetikzlibrary{positioning,shapes,shadows,arrows,calc}
\tikzstyle{component}=[rectangle, draw=black, rounded corners, fill=blue!40, drop shadow, text centered, anchor=north, text=white, minimum height=1cm]
\tikzstyle{arrow}=[->, thick]

\usetikzlibrary{intersections}

\usepackage{xcolor}  
\definecolor{myblue}{RGB}{34,31,217}

\definecolor{mycyan}{gray}{.7}

\newtheorem{theorem}{Theorem}

\newtheorem{corollary}{Corollary}

\newcommand{\pref}{\prettyref}

\newrefformat{fig}{Fig.~\ref{#1}}
\newrefformat{tab}{Table~\ref{#1}}
\newrefformat{sec}{Section~\ref{#1}}
\newrefformat{app}{Appendix~\ref{#1}}
\newrefformat{alg}{Algorithm~\ref{#1}}
\newrefformat{property}{Property~\ref{#1}}
\newrefformat{theorem}{Theorem~\ref{#1}}
\newrefformat{corollary}{Corollary~\ref{#1}}
\newrefformat{proposition}{Proposition~\ref{#1}}
\newrefformat{def}{Definition~\ref{#1}}
\newrefformat{eq}{equation~(\ref{#1})}

\usepackage{graphicx}
\definecolor{Gray}{gray}{0.9}
\usepackage{scrpage2}

\usepackage{lscape}

\begin{document}

\title{\textbf\LARGE\fontfamily{cmss}\selectfont Integration of Preferences in Decomposition Multi-Objective Optimization\footnote{This work has been submitted to the IEEE for possible publication. Copyright may be transferred without notice, after which this version may no longer be accessible.}}

\author[1]{\normalsize\fontfamily{lmss}\selectfont Ke Li}
\author[2]{\normalsize\fontfamily{lmss}\selectfont Kalyanmoy Deb}
\author[3]{\normalsize\fontfamily{lmss}\selectfont Xin Yao}
\affil[1]{\normalsize\fontfamily{lmss}\selectfont College of Engineering, Mathematics and Physical Sciences, University of Exeter}
\affil[2]{\normalsize\fontfamily{lmss}\selectfont Department of Electrical and Computer Engineering, Michigan State University}
\affil[3]{\normalsize\fontfamily{lmss}\selectfont CERCIA, School of Computer Science, University of Birmingham}
\affil[$\ast$]{\normalsize\fontfamily{lmss}\selectfont Email: k.li@exeter.ac.uk, kdeb@egr.msu.edu, x.yao@cs.bham.ac.uk}

\renewcommand\Authands{ and }

\date{}
\maketitle

{\normalsize\fontfamily{lmss}\selectfont\textbf{Abstract: } }Most existing studies on evolutionary multi-objective optimization focus on approximating the whole Pareto-optimal front. Nevertheless, rather than the whole front, which demands for too many points (especially in a high-dimensional space), the decision maker might only interest in a partial region, called the region of interest. In this case, solutions outside this region can be noisy to the decision making procedure. Even worse, there is no guarantee that we can find the preferred solutions when tackling problems with complicated properties or a large number of objectives. In this paper, we develop a systematic way to incorporate the decision maker's preference information into the decomposition-based evolutionary multi-objective optimization methods. Generally speaking, our basic idea is a non-uniform mapping scheme by which the originally uniformly distributed reference points on a canonical simplex can be mapped to the new positions close to the aspiration level vector specified by the decision maker. By these means, we are able to steer the search process towards the region of interest either directly or in an interactive manner and also handle a large number of objectives. In the meanwhile, the boundary solutions can be approximated given the decision maker's requirements. Furthermore, the extent of the region of the interest is intuitively understandable and controllable in a closed form. Extensive experiments, both proof-of-principle and on a variety of problems with 3 to 10 objectives, fully demonstrate the effectiveness of our proposed method for approximating the preferred solutions in the region of interest.

{\normalsize\fontfamily{lmss}\selectfont\textbf{Keywords: } }evolutionary multi-objective optimization, decomposition-based method, user-preference incorporation, reference points

\input{introduction}
\input{related}
\input{proposal}
\input{empirical}
\input{conclusion}

\bibliographystyle{IEEEtran}
\bibliography{IEEEabrv,preference}

\end{document}

%% file: introduction.tex

\section{Introduction}
\label{sec:introduction}

Most real-life applications usually consider optimizing multiple incommensurable and conflicting objectives simultaneously. To handle such problems, termed as multi-objective optimization problems (MOPs), decision makers (DMs) often look for a set of Pareto-optimal solutions none of which can be considered better than another when all objectives are of importance. Evolutionary multi-objective optimization (EMO) algorithms, which work with a population of solutions and can approximate a set of trade-off solutions simultaneously, have been widely accepted as a major tool for solving MOPs. Over the past two decades and beyond, many efforts have been devoted to developing EMO algorithms (e.g., elitist non-dominated sorting genetic algorithm (NSGA-II)~\cite{NSGA-II,SPEA2,ZhouZLL09,KeSMC12}, indicator-based EA (IBEA)~\cite{IBEA,BeumeNE07,BaderZ11,LiKCLZS12} and multi-objective EA based on decomposition (MOEA/D)~\cite{MOEAD,LiKZD15,LiFKZ14,LiKD15}) to find a set of efficient solutions that well approximate the whole Pareto-optimal front (PF) in terms of convergence and diversity.

The ultimate goal of multi-objective optimization is to help the DM find solutions that meet at most his/her preferences. Supplying a DM with a large amount of trade-off points, which approximate the whole PF, not only increases his/her workload, but also provides many irrelevant or even noisy information to the decision making procedure. Moreover, due to the curse of dimensionality, approximating a high-dimensional PF at a whole not only becomes computationally inefficient (or even infeasible), but also causes a severe cognitive obstacle for the DM to comprehend the high-dimensional data. To facilitate the decision making procedure, it is more practical to incorporate the DM's preference information into the search process. This allows the computational efforts to concentrate on the region of interest (ROI) and thus has a better approximation therein. In general, the preference information can be incorporated \textit{a priori}, \textit{posteriori} or \textit{interactively}. Note that the traditional EMO goes along the posteriori way of which the disadvantages have been described before. If the preference information is elucidated a priori, it is used to guide the solutions towards the ROI. However, it is non-trivial to faithfully model the preference information before solving the MOP at hand. In practice, articulating the preference information in an interactive manner, which has been studied in the multi-criterion decision making (MCDM) field for over half a century, seems to be more interesting. This enables the DMs to progressively learn and understand the characteristics of the MOP at hand and adjust their articulated preference information. As a consequence, the solutions are effectively driven towards the ROI.

Recently, there have been some initiatives on integrating and blending the EMO and MCDM together to tailor the DM's preference information~\cite{Coello00,RachmawatiS06,PurshouseDMMW14,BechikhKSG15}. Nevertheless, although the existing works aim at steering the search process towards the ROI, the definition of the ROI is still vague. First of all, the ROI can be any part of the PF near the DM specified aspiration level vector or even subjectively determined by the DM. Secondly, the ROI is expected to be a partial region of the PF whereas no quantitative definition has been given to the size of this region. Although some studies (e.g.,~\cite{DebSBC06,SaidBG10,ThieleMKL09}) claimed to control the spread of the preferred solutions accommodating to the DM's expectation of the extent of the ROI, i.e., the ROI's size, the corresponding parameter setting is ad-hoc~\cite{BechikhKSG15}. In addition to the ROI, the boundary of the PF is also important for the DM to understand the underlying problem and to facilitate the further decision making procedure. In particular, the boundary provides the DM a general information about the PF's geometrical characteristics; and more importantly, it provides the information of the ideal and nadir points which facilitate the normalization of the disparately scaled objective functions. But unfortunately, to the best of our knowledge, how to keep the solutions located in the ROI and the boundary simultaneously has rarely been studied yet.

During recent years, especially after the developments of MOEA/D and NSGA-III~\cite{DebJ14}, the decomposition-based EMO methods have become increasingly popular for the posteriori multi-objective optimization. Generally speaking, by specifying a set of reference points\footnote{In this paper, we use the term reference point without loss of generality, although some other literatures, e.g., the original MOEA/D~\cite{MOEAD}, also use the term weight vector interchangeably.}, the decomposition-based EMO methods at first decompose the MOP at hand into multiple subproblems, either with scalar objective or simplified multi-objective. Then, a population-based technique is applied to solve these subproblems in a collaborative manner. Under some mild conditions, the optimal solutions of all subproblems constitute a good approximation to the PF. It is not difficult to understand that the distribution of the reference points is critical in a decomposition-based EMO method. It not only implies a priori prediction of the PF's geometrical characteristics, but also determines the distribution and uniformity of optimal solutions. There have been some studies on how to generate uniformly distributed reference points. For example, \cite{NBI} and \cite{TanJLW12} suggested some structured methods to generate uniformly distributed reference points on a canonical simplex. To adapt to the irregular PFs, such as disconnected or mixed shapes and disparately scaled objectives, some adaptive reference point adjustment methods (e.g., \cite{JiangCZO11} and \cite{QiMLJSW14}) have been developed to progressively adjust the distribution of reference points on the fly. To integrate the DM's preference information into the decomposition-based EMO methods, a natural idea is to bias the distribution of the reference points towards the ROI. Although it sounds quite intuitive, in practice, how to obtain the appropriate reference points that accommodate to the DM's preference information is far from trivial. Most recently, there have been some limited initiatives on adjusting the distribution of the reference points according to the DM's preference information (e.g.,~\cite{GiagkiozisF14,MaLQLJDWDHZW15,ChengJOS16}). However, they are ad-hoc and the position and extent of the reference points around the ROI are not fully controllable.

In this paper, we present a systematic way to incorporate the DM's preference information, either a priori or interactively, into the decomposition-based EMO methods. In particular, here we model the DM's preference information as an aspiration level vector, which has been widely used in this literature~\cite{BechikhKSG15}. More specifically, our basic idea is a non-uniform mapping scheme by which the uniformly distributed reference points on a canonical simplex can be mapped to the new positions close to the DM specified aspiration level vector and thereby having a biased distribution. The mapping function is nonlinear in nature and is a function of a reference point's position with respect to the pivot point. In particular, this pivot point is the representative of the ROI on the simplex and determines the ROI's position. There are three major properties of our proposed non-uniform mapping scheme:
\begin{itemize}
    \item The distribution of the reference points after the non-uniform mapping is biased towards the pivot point. In other words, the closer to the pivot point, the more reference points since they are more relevant to the DM's preference information.
    \item The extent of the biased reference points is fully controllable. In particular, we provide an intuitively understandable definition to quantify the extent of the ROI, which is a quantity proportional to the area of the whole PF.
    \item The reference points after the non-uniform mapping not only have a biased distribution towards the ROI, but also the ones located on the boundary are able to be preserved in the meanwhile. This latter property enables a decomposition-based EMO method not only find the preferred solutions, but also provide the global information about the PF to the DM.
\end{itemize}

The rest of this paper is organized as follows. \pref{sec:relatedworks} devotes to overviewing some state-of-the-art related to this paper. \pref{sec:proposal} presents the technical details of our proposed non-uniform mapping scheme. \pref{sec:proofofprinciple} and \pref{sec:comparisons} show the empirical studies and analysis on a series of benchmark problems. Finally, \pref{sec:conclusion} provides some concluding remarks along with some pertinent observations.

%% file: related.tex

\section{Related Works}
\label{sec:relatedworks}

\begin{table*}[htbp]
\scriptsize
\centering
\caption{Comparison of Preference-based EMO Methodologies (Inspired by~\cite{BechikhKSG15})}
\label{tab:preference-tool}
\begin{tabular}{c|c|c|c|c|c|c}
\hline
Preference Incorporate                   & Publications                                   & Modifcation               & Influence    & Multiple ROIs & ROI Control & Scalability \\\hline
\multirow{2}{*}{Weight}                  & \cite{Deb03}                                   & density estimation        & distribution & $\times$      & $\times$    & $\times$    \\\cline{2-7}
                                         & \cite{JinOS01}                                 & objective aggregation     & distribution & $\times$      & $\times$    & $\times$    \\\hline
\multirow{6}{*}{Trade-off information}   & \cite{GreenwoodHD96}                           & dominance                 & region       & $\times$      & $\times$    & $\times$    \\\cline{2-7}
                                         & \cite{DebSKW10}                                & dominance                 & region       & $\times$      & $\times$    & \checkmark  \\\cline{2-7}
                                         & \cite{BattitiP10}                              & density estimation        & region       & $\times$      & $\times$    & \checkmark  \\\cline{2-7}
                                         & \cite{Branke2001499}                           & dominance                 & region       & $\times$      & $\times$    & $\times$    \\\cline{2-7}
                                         & \cite{FowlerGKKMW10}                           & dominance                 & region       & $\times$      & $\times$    & $\times$    \\\cline{2-7}
                                         & \cite{ParmeeC02}                               & dominance                 & distribution & $\times$      & \checkmark  & $\times$    \\\hline
\multirow{2}{*}{Density}                 & \cite{BrankeD05}                               & density estimation        & distribution & $\times$      & \checkmark  & $\times$    \\\cline{2-7}
                                         & \cite{KoksalanK10}                             & dominance                 & distribution & $\times$      & \checkmark  & $\times$    \\\hline
\multirow{4}{*}{Indicator}               & \cite{ThieleMKL09}                             & quality indicator         & region       & \checkmark    & \checkmark  & \checkmark  \\\cline{2-7}
                                         & \cite{WagnerT10}                               & objective functions       & region       & \checkmark    & $\times$    & \checkmark  \\\cline{2-7}
                                         & \cite{TrautmannWB13}                           & quality indicator         & distribution & $\times$      & $\times$    & \checkmark  \\\cline{2-7}
                                         & \cite{WagnerTB13}                              & quality indicator         & region       & $\times$      & \checkmark  & \checkmark  \\\hline
\multirow{6}{*}{Aspiration level vector} & \cite{FonsecaF98}                              & dominance                 & region       & $\times$      & $\times$    & $\times$    \\\cline{2-7}
                                         & \cite{WickramasingheL08,AllmendingerLB08}      & leader selection strategy & region       & \checkmark    & \checkmark  & $\times$    \\\cline{2-7}
                                         & \cite{DebSBC06,DebK07,DebK07CEC}               & density estimation        & region       & \checkmark    & \checkmark  & \checkmark  \\\cline{2-7}
                                         & \cite{MolinaSDCC09}                            & dominance                 & region       & \checkmark    & $\times$    & $\times$    \\\cline{2-7}
                                         & \cite{r-dominance}                             & dominance                 & region       & \checkmark    & \checkmark  & \checkmark  \\\cline{2-7}
                                         & \cite{ChengJOS16,Yu2015,Ma2015,MohammadiOLD14} & weight vector             & region       & \checkmark    & \checkmark  & \checkmark  \\\hline
\end{tabular}

\begin{tablenotes}
\item[1] \textit{Modification} indicates the modified part of the EMO algorithm. \textit{Influence} indicates whether the result is bounded ROI or a biased distribution. \textit{Scalability} indicates whether it is scalable to any number of objectives. \textit{ROI control} indicates whether the extent of the ROI is controllable.
\end{tablenotes}
\end{table*}

During the past decades, various methods have been developed to incorporate the DM's preference information into the EMO~\cite{Coello00,RachmawatiS06,PurshouseDMMW14,BechikhKSG15}. In this section, we briefly overview the existing literature according to the way of articulating the DM's preference information, as summarized in~\pref{tab:preference-tool}.

The first one employs the weight information, i.e., relative importance, to model the DM's preference information. For example, \cite{Deb03} developed a modified fitness sharing mechanism, by using a weighted Euclidean distance, to bias the population distribution. \cite{JinOS01} developed a method to convert the fuzzy linguistic preference information into an interval-based weighting scheme where the weights is perturbed between the pre-defined upper and lower bounds. By transforming the MOP into a series of single-objective aggregation functions, e.g., weighted sum, it guides the population towards the ROI. It is worth noting that the weight-based methods become ineffective when facing a large number of objectives. Because it is difficult to either specify the weights or verify the quality of the biased approximation. Moreover, it is unintuitive and challenging for the DM to steer the search process towards the ROI via the weighting scheme. In addition, the weight-based methods are unable to approximate multiple ROIs and control the extent of the ROI.

The second sort modifies the trade-off information by either classifying objectives into different levels and priorities or expressing the DM's preference information via the fuzzy linguistic terms according to different aspiration levels. For example, \cite{GreenwoodHD96} suggested an imprecisely specified multi-attribute utility theory-based weighted sum method to obtain the ranking of objectives from some candidate solutions. In~\cite{DebSKW10} and~\cite{BattitiP10}, the authors employed the concept of value function to design the preference-based EMO algorithms. In particular, the precise form of the value function is unknown a priori, whereas it is progressively learnt from the optimization process by the interaction with the DM. In~\cite{Branke2001499} and~\cite{FowlerGKKMW10}, the DM's preference information is used to develop some modified trade-off relationship to compare solutions. \cite{ParmeeC02} suggested a method to integrate the DM's fuzzy preference information into the EMO algorithm by converting the linguistic terms into weights. This sort of methods is interesting but complicated, especially when the number of objectives becomes large~\cite{SaidBG10}. In addition, using such an approach interactively increases the DM's burden and it is hard to control the extent of the ROI.

The third category tries to bias the density of solutions towards the ROI by considering the DM's preference information. \cite{BrankeD05} developed a mapping method to modify the crowding distance calculation in NSGA-II by which the search process can be guided towards the ROI. In~\cite{KoksalanK10}, the biased distribution of the solutions is achieved by setting different territory sizes in the territory-based evolutionary algorithm~\cite{KarahanK10}. In particular, a smaller territory leads to a higher resolution of solutions, and vice versa. The major drawback of this sort of methods comes from the diversity management itself, especially in a high-dimensional space. Due to the same reason, it cannot control the extent of the ROI precisely.

The fourth class, as a recent trend, combines the DM's preference information with the performance indicator in the algorithm design. For example, \cite{ThieleMKL09} employed a modified binary quality indicator, which incorporates the DM's preference information, to assign the fitness value to each solution. In \cite{WagnerT10}, the objective functions of the original MOP are at first converted into the desirability functions (DFs). Afterwards, a popular indicator-based EMO algorithm, i.e., SMS-EMOA~\cite{BeumeNE07}, is used as the search engine to approximate the ROI. In particular, the calculation of hypervolume is according to the DFs instead of the original objective functions. \cite{ZitzlerBT06} proposed a weighting scheme for the hypervolume indicator to guide the search process towards the ROI. In particular, the DM's preference information is articulated as the weighting coefficients and an aspiration level vector. In~\cite{TrautmannWB13} and~\cite{WagnerTB13}, the DM's preference information is integrated into the R2 indicator, a set-based performance indicator. Then, it is used to guide an indicator-based EMO algorithm to approximate the ROI. One of the major drawbacks of this sort of methods is the high computational cost for calculating the indicator, e.g., hypervolume~\cite{Abbz2012a}, which increases exponentially with the number of objectives. In addition, although they claimed that the extent of the ROI is controllable by some specific parameters. Unfortunately, there is no rule-of-thumb for specifying the appropriate parameters that accommodate the DM's expectations of the ROI's size.

The last one uses the aspiration level vectors, which represent the DM's desired values for each objective, to steer the search process. As the first attempt to incorporate the DM's preference information in EMO, \cite{FonsecaF98} suggested to model the DM's preference as a goal, i.e., the aspiration level vector, to achieve. In~\cite{DebSBC06,DebK07} and~\cite{DebK07CEC}, the authors combined the reference point, i.e., aspiration level vector, related methods with NSGA-II to guide the search process towards the ROI. In particular, solutions close to the given reference point have a high priority to survive to the next generation. In~\cite{WickramasingheL08} and~\cite{AllmendingerLB08}, the aspiration level vector is used to help select the leader swarm in the multi-objective particle swarm optimization algorithm. \cite{MolinaSDCC09} suggested a modified dominance relationship, called g-dominance, where solutions satisfy either all or none aspiration levels are preferred over those satisfying some aspiration levels. \cite{r-dominance} developed another modified dominance relationship, called r-dominance, where non-dominated solutions, according to the Pareto dominance relationship, can be distinguished by their weighted Euclidean distances towards the DM supplied aspiration level vector. Recently, some decomposition-based methods also used the aspiration level vector to incorporate the DM's preference information into the search process, e.g.,~\cite{ChengJOS16} and~\cite{Yu2015,Ma2015,MohammadiOLD14}. Generally speaking, their basic idea is to use the aspiration level vector as the anchor point around which they try to obtain some reference points. Comparing to the other preference modeling tools, the aspiration level vector is the most natural and intuitive way to elucidate the DM's preference information. Without a demanding effort, the DM is able to guide the search towards the ROI directly or interactively even when encountering a large number of objectives. Nevertheless, the existing methods cannot approximate the solutions in the ROI and the boundary simultaneously. In addition, the control of the extent of the ROI is ad-hoc.

%% file: proposal.tex

\section{Non-Uniform Mapping Scheme}
\label{sec:proposal}

\subsection{Overview}
\label{sec:overview}

\begin{figure*}[htbp]
\centering

\pgfplotsset{every axis/.append style={
grid = major,
view   = {135}{30},
xlabel = {$f_1$},
ylabel = {$f_2$},
zlabel = {$f_3$},
xmin = 0, xmax = 1.005,
ymin = 0, ymax = 1.005,
zmin = 0, zmax = 1.005,
thick,
line width = 1pt,
tick style = {line width = 0.8pt}}}

  \subfloat[Uniform distribution]{
  \resizebox{0.33\textwidth}{!}{
      \begin{tikzpicture}
	\begin{axis}
	\addplot3+[only marks, black, mark options = {fill = black}, mark = *] table {Data/W3D_91.dat};
	\end{axis}
      \end{tikzpicture}
    }}
  \subfloat[Biased distribution with boundary]{
  \resizebox{0.33\textwidth}{!}{
      \begin{tikzpicture}
	\begin{axis}
	\addplot3+[only marks, black, mark options = {fill = black}, mark = *] table {Data/pW3D_DTLZ1_1.dat};
	\end{axis}
      \end{tikzpicture}
    }}
    \subfloat[Biased distribution without boundary]{
    \resizebox{0.33\textwidth}{!}{
      \begin{tikzpicture}
	\begin{axis}
	\addplot3+[only marks, black, mark options = {fill = black}, mark = *] table {Data/pW3D_DTLZ1_2.dat};
	\end{axis}
      \end{tikzpicture}
    }}
\caption{Reference points used in decomposition-based EMO methods.}
\label{fig:OverviewExample}
\end{figure*}
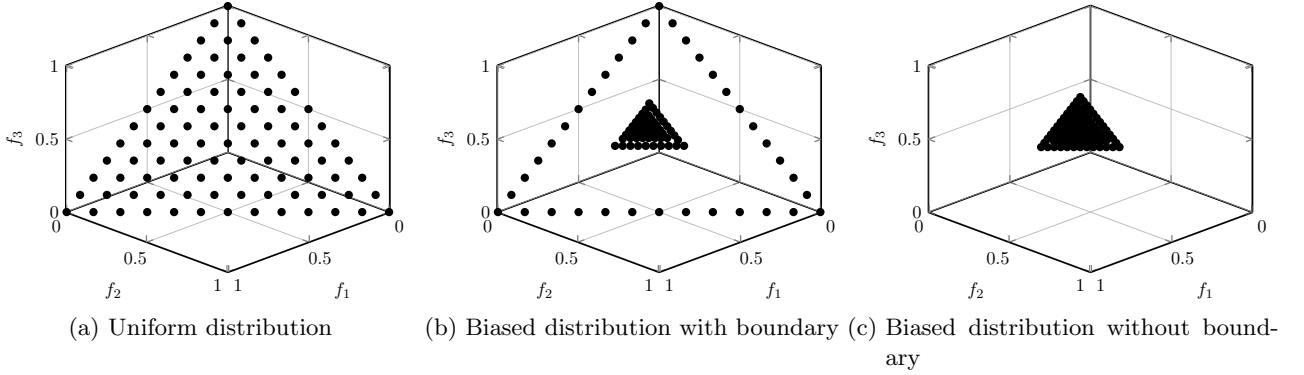

Reference points, as the basic components in the decomposition-based EMO algorithms, are usually generated in a structured manner, e.g., the Das and Dennis's method\footnote{In~\cite{NBI}, $N={H+m-1 \choose m-1}$ reference points, with a uniform spacing $\delta=\frac{1}{H}$, are sampled from a canonical simplex $\Psi^m$, where $H>0$ is the number of divisions considered along each objective coordinate, and $m$ is the number of objectives.}~\cite{NBI}. \pref{fig:OverviewExample}(a) shows an example of 91 uniformly distributed reference points in a three-dimensional space. In this case, the DM has no preference on any particular region of the PF. These reference points are used to guide a decomposition-based EMO algorithm search for the whole PF. On the other hand, if the DM has elucidated some preference information, e.g., an aspiration level vector, it is preferable that reference points can have a biased distribution towards the ROI accordingly. Bearing this consideration in mind, this section presents a non-uniform mapping scheme (NUMS) by which we are able to bias the distribution of reference points, originally generated in a structured manner, towards the ROI. \pref{fig:OverviewExample}(b) and \pref{fig:OverviewExample}(c) show two examples of the biased reference points distribution after the non-uniform mapping. In the following paragraphs, we will describe the mathematical model of NUMS in detail before showing its algorithmic implementations.

\subsection{Non-Uniform Mapping Scheme in One-Dimensional Space}
\label{sec:mappingmodel}

\begin{figure}[htbp]
\centering
\includegraphics[width=.5\linewidth]{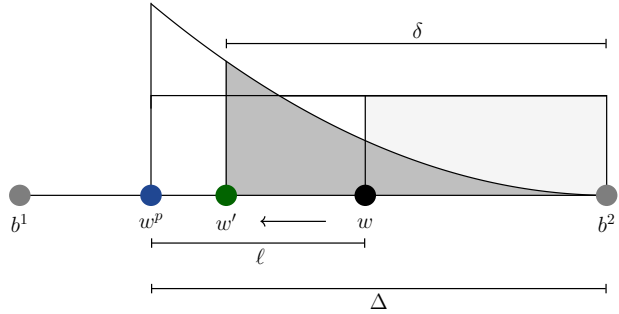}
\caption{An example of boundary intersection in three dimensional space.}
\label{fig:mapping}
\end{figure}

Let us begin with describing the mathematical model of NUMS in a one-dimensional case. Considering the illustrative example shown in~\pref{fig:mapping}, the reference points generated in a structured manner are uniformly distributed along the line starting from $b^1$ and ending at $b^2$. Let us assume that the position of a uniformly distributed reference point $w$ obeys a uniform distribution whose probability density function (PDF) is defined as follows:
\begin{equation}
\psi^u(\zeta)=\frac{1}{\Delta}
\label{eq:uniform}
\end{equation}

where $0\leq\zeta\leq\Delta$, $\Delta=|b^2-w^p|$ is the distance between between $w^p$ and $b^2$. Here $w^p$ is defined as the pivot point, which will be further discussed in Appendix 1 of the supplementary file\footnote{The supplementary file can be downloaded at https://coda-group.github.io/publications/suppNUMS.pdf}, to represent the ROI. When considering the DM's preference information, rather than a uniform distribution, it is preferable that the reference points have a biased distribution towards $w^p$, i.e., the closer to $\mathbf{w}^p$, the more reference points. The purpose of NUMS is to shift $w$, originally generated by a structured manner, onto a new position $w'$ close to $w^p$. Let us assume that the position of $w'$ obeys an exponential distribution whose PDF is defined as follows:
\begin{equation}
\psi^e(\xi)=k\xi^{\eta}
\label{eq:mapping1}
\end{equation}
where $\xi=\frac{\delta}{\Delta}$, $\delta=|b^2-w'|$ is the distance between $w'$ and $b^2$, and $\eta$ is a control parameter which will be further discussed in~\pref{sec:etasetting}. Notice that $0\leq\xi\leq 1$ and $\delta$ gives the exact position of $w'$ along the line starting from $b^1$ and ending at $b^2$. By equating the area under the probability curve of $\psi^e(\xi)$ with that of $\psi^u(\zeta)$, we have:
\begin{equation}
\int_0^{\frac{\delta}{\Delta}}kx^{\eta}dx=\int_0^{\Delta-\ell}\frac{1}{\Delta}dx=\frac{\Delta-\ell}{\Delta}
\label{eq:mapping2}
\end{equation}
where $\ell=|w-w^p|$ is the distance between $w$ and $w^p$. By letting $\ell=0$ and $\delta=\Delta$ in~\pref{eq:mapping2}, we have:
\begin{equation}
\int_0^1kx^{\eta}dx=1
\label{eq:mapping3}
\end{equation}
this gives us $k=\eta+1$. Finally, by substituting $\eta+1$ for $k$ in~\pref{eq:mapping2}, we have:
\begin{equation}
\delta=\Delta(\frac{\Delta-\ell}{\Delta})^{\frac{1}{\eta+1}}
\label{eq:delta}
\end{equation}

\subsection{Vector-Wise Mapping in $m$-Dimensional Space}
\label{sec:vectormapping}

\begin{figure}[htbp]
\centering
\begin{tikzpicture}

\begin{axis}[
	thick,
	width = 7cm,
	height = 7cm,
   	xmin = 0,
  	xmax = 1.1,
  	ymin = 0,
  	ymax = 1.1, 
  	axis lines = left,
	enlargelimits = 0.0
  ]
	\addplot[smooth, thick, name path=A] table {Data/data2D.dat};
	\addplot[smooth, thick] table {Data/dataLine.dat};
	\addplot[smooth, thick] table {Data/dataLine2.dat};
	\addplot[thick] table {Data/dataLine3.dat};
	\addplot[thick] table {Data/dataLine4.dat};
	\addplot[thick] table {Data/dataLine5.dat};
	
\end{axis}
  \fill[fill = black!50] (0.0, 4.9) circle (4pt);
	\fill[fill = {rgb:red,1;green,2;blue,4}] (1.43, 3.46) circle (4pt);
	\node[below] (P) at (1.15, 3.45) {$\mathbf{w}^p$};
	\fill[fill = black!60!green] (2.05, 2.85) circle (4pt);
	\node[below] (P) at (1.75, 2.85) {$\mathbf{w}'$};
	\fill (3.17, 1.72) circle (4pt);
	\node[below] (P) at (2.9, 1.7) {$\mathbf{w}$};
	\fill[fill = black!50] (4.9, 0.0) circle (4pt);
	\node[below] (P) at (5.3, 0.0) {$\mathbf{b}$};
	
	\draw[<-, thick] (2.05, 2.4) -- (2.7, 1.75);
	\draw[|-|, thick] (0.9, 2.95) -- (2.6, 1.25);
    \node[below] (sd) at (1.5, 2.2) {$\ell$};
    \draw[|-|, thick] (0.5, 2.6) -- (4.1, -1);
    \node[below] (sd) at (2.0, 1.05) {$\Delta$};
    \draw[|-|, thick] (2.88, 3.67) -- (5.8, 0.8);
    \node[above] (sd) at (4.5, 2.2) {$\delta$};

\end{tikzpicture}
\caption{Non-uniform mapping scheme in 2-D scenario.}
\label{fig:mapping2D}
\end{figure}

Now, we generalize the one-dimensional non-uniform mapping model into a $m$-dimensional case. Without loss of generality, let us consider a two-dimensional example shown in~\pref{fig:mapping} for illustration. Similar to the one-dimensional case, the purpose of NUMS in a $m$-dimensional case is to shift an uniformly distributed reference point $\mathbf{w}$ onto $\mathbf{w}'$ along the direction $\mathbf{w}^p-\mathbf{w}$. For the ease of latter computation, we consider in an opposite direction. That is to say NUMS shifts $\mathbf{w}^p$ onto $\mathbf{w}'$ along the direction $\mathbf{w}-\mathbf{w}^p$. Accordingly, $\mathbf{w}'$ is calculated as:
\begin{equation}
\mathbf{w}'=\mathbf{w}^p+\rho\times\frac{\mathbf{w}-\mathbf{w}^p}{\|\mathbf{w}-\mathbf{w}^p\|}
\label{eq:newposition}
\end{equation}
where $\|\cdot\|$ represents $\ell^2$-norm and $\rho$ is calculated as:
\begin{equation}
\rho=\Delta-\delta
\label{eq:upsilon}
\end{equation}
where $\Delta=\|\mathbf{b}-\mathbf{w}^p\|$ and $\delta$ is calculated based on~\pref{eq:delta} in which $\ell=\|\mathbf{w}-\mathbf{w}^p\|$. Notice that $\mathbf{w}$ and $\mathbf{w}^p$ are known a priori, while $\mathbf{b}$ is one of the intersecting points between the line connecting $\mathbf{w}^p$ and $\mathbf{w}$ and the edges of the simplex $\Psi^m$. Generally speaking, $\mathbf{b}$ can be calculated as:
\begin{equation}
\mathbf{b}=\mathbf{w}^p+\Delta\times\frac{\mathbf{w}-\mathbf{w}^p}{\|\mathbf{w}-\mathbf{w}^p\|}
\label{eq:boundary}
\end{equation}
Geometrically, there are at most $m$ such intersecting points, each of which should have a zero element. In this case, for each $\mathbf{b}^i$, $i\in\{1,\cdots,m\}$, the corresponding $\Delta$ in~\pref{eq:boundary} can be calculated as:
\begin{equation}
\Delta=\min\limits_{1\leq i\leq m}[w^p_i\times\frac{\|\mathbf{w}^p-\mathbf{w}\|}{w^p_i-w_i}]_+
\label{eq:deltacomputation}
\end{equation}
where $[\sigma]_+$ returns $\sigma$ if and only if $\sigma>0$, otherwise it returns an invalid number.




\subsection{Setting of $\eta$}
\label{sec:etasetting}

\begin{figure}[htbp]
\centering
\begin{tikzpicture}
\begin{axis}[
		grid   = major,
		xmin   = 0, xmax = 1,
		ymin   = 0, ymax = 12,
		xlabel = {$\xi$},
		ylabel = {$\psi^e(\xi)$},
        legend entries = {$\eta=-0.5$,$\eta=0.5$,$\eta=1.0$,$\eta=2.0$,$\eta=5.0$,$\eta=10.0$},
        legend style = {nodes = right},
		line width = 1pt,
		tick style = {line width = 0.8pt}
	]
  \addplot[brown, mark = +] table {Data/eta_neg05.dat};
	\addplot[black, mark = x] table {Data/eta_05.dat};
	\addplot[cyan, mark = diamond] table {Data/eta_1.dat};
	\addplot[green, mark = triangle] table {Data/eta_2.dat};
	\addplot[red, mark = o] table {Data/eta_5.dat};
	\addplot[blue, mark = square] table {Data/eta_10.dat};
\end{axis}
\end{tikzpicture}
\caption{Non-uniform mapping function with different $\eta$.}
\label{fig:eta}
\end{figure}
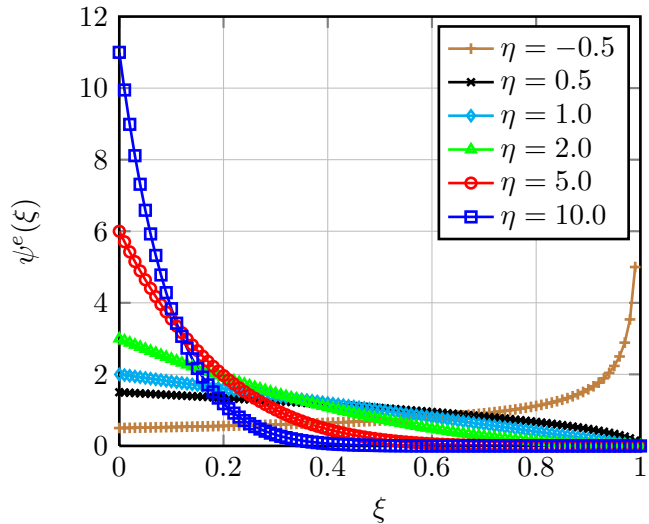

In~\pref{eq:mapping1}, $\eta$ controls the gradient of the PDF curve. \pref{fig:eta} shows six function curves with various $\eta$ settings. More specifically, $\psi^e(\xi)$ is a decreasing function of $\xi$ when $\eta>0$; while it is an increasing function of $\xi$ when $\eta<0$. From~\pref{fig:eta}, we also find that the function curve is more skewed with a larger $\eta$. According to the properties of power function, it is not difficult to understand that, for a given $\Delta$ and $\ell$ in~\pref{eq:delta}, a larger $\eta$ will results in a larger $\delta$. In summary, $\eta$ has the following two effects on the NUMS:
\begin{itemize}
\item To push $\mathbf{w}$ towards $\mathbf{w}^p$, we need to set $\eta>0$; otherwise $\mathbf{w}$ will be shifted away from $\mathbf{w}^p$. 
\item With a large $\eta$, which results in a large $\delta$, $\mathbf{w}'$ has a large probability to be closer to $\mathbf{w}^p$ after non-uniform mapping; on the flip side, $\mathbf{w}'$ will be more probably closer to $\mathbf{b}$.
\end{itemize}

Based on the above discussions, we realize that $\eta$ is able to control the extent of the biased reference points after the NUMS. However, due to the non-linear property of the PDF in~\pref{eq:mapping1}, it is far from trivial to choose the appropriate $\eta$ beforehand that results in the expected extent of the ROI. Instead of understanding the non-linear mapping function, the following theorem provides a closed form method for setting the appropriate $\eta$ value.
\begin{theorem}
Suppose the relative extent of reference points after non-uniform mapping is $\tau$ $(0<\tau\leq 1)$, comparing to the simplex $\Psi^m$, the $\eta$ value in~\pref{eq:mapping1} is calculated as
\begin{equation}
\eta = \frac{\log\alpha}{\log\beta} - 1
\end{equation}
where $\alpha=\frac{m}{H}$ and $\beta=1-\tau$.
\label{theorem:eta}
\end{theorem}

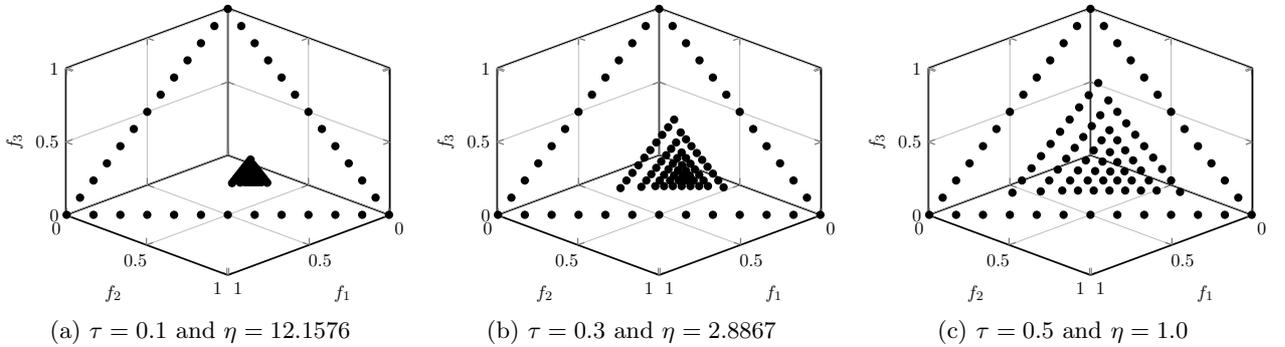
\begin{figure*}[htbp]

\pgfplotsset{every axis/.append style={
grid = major,
view   = {135}{30},
xlabel = {$f_1$},
ylabel = {$f_2$},
zlabel = {$f_3$},
xmin = 0, xmax = 1.005,
ymin = 0, ymax = 1.005,
zmin = 0, zmax = 1.005,
thick,
line width = 1pt,
tick style = {line width = 0.8pt}}}

  \subfloat[$\tau=0.1$ and $\eta=12.1576$]{
  \resizebox{0.33\textwidth}{!}{
      \begin{tikzpicture}
	\begin{axis}
	\addplot3+[only marks, black, mark options = {fill = black}, mark = *] table {Data/eta1.dat};
	\end{axis}
      \end{tikzpicture}
    }}
  \subfloat[$\tau=0.3$ and $\eta=2.8867$]{
  \resizebox{0.33\textwidth}{!}{
      \begin{tikzpicture}
	\begin{axis}
	\addplot3+[only marks, black, mark options = {fill = black}, mark = *] table {Data/eta3.dat};
	
	\end{axis}
      \end{tikzpicture}
    }}
    \subfloat[$\tau=0.5$ and $\eta=1.0$]{
    \resizebox{0.33\textwidth}{!}{
      \begin{tikzpicture}
	\begin{axis}
	\addplot3+[only marks, black, mark options = {fill = black}, mark = *] table {Data/eta5.dat};
	\end{axis}
      \end{tikzpicture}
    }}
\caption{Distribution of reference points for different settings of $\tau$ and their corresponding $\eta$ when $\mathbf{z}^r=(0.7,0.8,0.5)^T$}
\label{fig:differenttau}
\end{figure*}

The proof of~\pref{theorem:eta} can be found in Appendix 2 of the supplementary file. In practice, rather than a concrete extent of the ROI, it is more plausible for the DM to specify a relative quantity. Here we use $\tau$, the ratio of the surface area of the ROI proportional to the whole PF, as this quantity. \pref{fig:differenttau} shows three examples of biased reference points after the non-uniform mapping with different $\tau$ settings. Based on~\pref{theorem:eta}, we have the following corollary which provides the upper and lower bounds for setting $\tau$.
\begin{corollary}
To make the extent of the adapted reference points shrink, we need set $0<\tau<1-\frac{m}{H}$.
\label{corollary:boundseta}
\end{corollary}

The proof of~\pref{corollary:boundseta} can be found in Appendix 3 of the supplementary file. In principle, comparing to the whole PF, the relative extent of the ROI can be any number between 0 and 1. However, \pref{corollary:boundseta} provides a restriction on $\tau$ in order to make the uniformly distributed reference points shrink to the ROI; otherwise they will expand towards the boundary. It is worth noting that \pref{theorem:eta} and \pref{corollary:boundseta} are derived under the condition $H>m$. Otherwise, all reference points generated by the Das and Dennis's method should lie on the boundary of the simplex $\Psi^m$. How to shift the reference points lying on the boundary will be described in the next subsection.

\subsection{Boundary Preservation}
\label{sec:boundary}

Notice that the NUMS described so far shifts the reference points, except those lying on the boundaries of the simplex $\Psi^m$, onto the ROI. The adapted reference points try to guide a decomposition-based EMO algorithm not only search for the preferred solutions, but also approximate those lying on the PF's boundaries. In particular, the boundary solutions provide the DM more comprehension of the PF, e.g., the PF's general shape, the ideal and nadir points which can be useful for further decision making. Nevertheless, if the DM does not interest in the boundary any longer, we can make a simple modification on the NUMS to shift the reference points lying on the boundaries towards the ROI as well. Specifically, a reference point $\mathbf{w}^b$ is considered lying on the boundary of $\Psi^m$ if and only if the following condition is met:
\begin{equation}
\Delta-\|\mathbf{w}^b-\mathbf{w}^p\|<\epsilon
\end{equation}
where $\epsilon=10^{-6}$ is a small quantity and $\Delta$ is determined according to~\pref{eq:deltacomputation}. To shift $\mathbf{w}^b$ onto the ROI, its new position after the NUMS is calculated as:
\begin{equation}
\mathbf{w}'=\mathbf{w}^p+\rho\times\frac{\mathbf{w}^b-\mathbf{w}^p}{\|\mathbf{w}^b-\mathbf{w}^p\|}
\label{eq:boundaryshift}
\end{equation}
where $\rho=\tau\times\|\mathbf{w}^b-\mathbf{w}^p\|$. Notice that the $\eta$ value obtained in \pref{theorem:eta} is derived in the situation that the DM is willing to keep the boundary points. If the reference points lying on the boundary are shifted onto the ROI by the NUMS as well,  the $\eta$ value should be calculated according to \pref{corollary:etaboundary}

\begin{corollary}
If all reference points are shifted onto the ROI, the $\eta$ value in \pref{eq:mapping1} is calculated as:
\begin{equation}
\eta=\frac{log\alpha}{log\beta}-1
\end{equation}
where $\alpha=\frac{m}{H}$ and $\beta=1-(1-\frac{m}{H})\times\tau$.
\label{corollary:etaboundary}
\end{corollary}

The proof of \pref{corollary:etaboundary} can be found in Appendix 4 of the supplementary file. Accordingly, we should have a different upper and lower bounds for $\eta$ as follows.
\begin{corollary}
If all reference points are shifted onto the ROI, we can set $0<\tau<1$.
\label{corollary:newbound}
\end{corollary}

The proof of \pref{corollary:newbound} can be found in Appendix 5 of the supplementary file. \pref{fig:OverviewExample}(c) gives an example that all reference points have been shifted onto the ROI.

\begin{algorithm}

\KwIn{
\begin{itemize}
	\item DM supplied aspiration level vector $\mathbf{z}^r$
	\item Number of divisions $H$
	\item Expected extent of ROI $\tau$
  \item \textsf{flag} determines whether keep the boundary or not
\end{itemize}
}
\KwOut{
\begin{itemize}
	\item Biased reference points $\overline{W}\leftarrow\{\overline{\mathbf{w}}^1,\cdots,\overline{\mathbf{w}}^N\}$
\end{itemize}
}

Initialize $N\leftarrow{H+m-1 \choose m-1}$ reference points $\mathbf{w}^1,\cdots,\mathbf{w}^N$ on a canonical simplex $\Psi^m$ by Das and Dennis's method;\\

Find the pivot point $\mathbf{w}^p$ of $\mathbf{z}^r$ on $\Psi^m$;\\

\uIf(\tcp*[f]{keep the boundary}){$\textsf{flag}=1$}{
$\alpha\leftarrow \frac{m}{H}$, $\beta\leftarrow 1-\tau$;
}\Else{
$\alpha\leftarrow \frac{m}{H}$, $\beta\leftarrow 1-(1-\frac{m}{H})\times\tau$;
}
$\eta\leftarrow\frac{\log\alpha}{\log\beta}-1$;\\

\For{$i\leftarrow 1$ \KwTo $N$}{
  $\Delta\leftarrow\min\limits_{1\leq j\leq m}[w_j^p\times\frac{\|\mathbf{w}^p-\mathbf{w}^i\|}{w_j^p-w_j^i}]_+$;\\
  \uIf{$\Delta-\|\mathbf{w}^i-\mathbf{w}^p\|<\epsilon\land\textsf{flag}=0$}{
    $\rho\leftarrow\tau\times\|\mathbf{w}^i-\mathbf{w}^p\|$;\\
  }\Else{
    $\delta\leftarrow\Delta(\frac{\Delta-\ell}{\Delta})^{\frac{1}{\eta+1}}$, where $\ell\leftarrow\|\mathbf{w}^i-\mathbf{w}^p\|$;\\
    $\rho\leftarrow\Delta-\delta$;\\
  }
  $\overline{\mathbf{w}}^i\leftarrow\mathbf{w}^p+\rho\times\frac{\mathbf{w}^i-\mathbf{w}^p}{\|\mathbf{w}^i-\mathbf{w}^p\|}$;\\
}
\Return $\overline{W}$

\caption{Non-uniform Mapping Scheme}
\label{alg:NUMS}
\end{algorithm}

\subsection{Algorithmic Details}
\label{sec:algorithmic}

After describing the mathematical foundations of the NUMS, this section gives its algorithmic details step by step. The pseudo-code of the NUMS is presented in~\pref{alg:NUMS}. First of all, $N={H+m-1 \choose m-1}$ reference points $\mathbf{w}^1,\cdots,\mathbf{w}^N$ are initialized via the Das and Dennis's method (line 1 of~\pref{alg:NUMS}). Afterwards, we use the method developed in~\cite{ChenY11} to find the pivot point, i.e., the projection of $\mathbf{z}^r$ on $\Psi^m$ (line 2 of~\pref{alg:NUMS} and more detailed discussions can be found in Appendix 1). Then, if the DM is interested in the boundary, we use \pref{theorem:eta} to compute the exponent $\eta$ of the PDF in~\pref{eq:mapping1}; otherwise, we use \pref{corollary:etaboundary} to do so (line 3 to line 7 of~\pref{alg:NUMS}). During the main loop, for each reference point, we use \pref{eq:deltacomputation} to determine the position of the corresponding boundary point for the non-uniform mapping (line 9 of \pref{alg:NUMS}). If the current investigating reference point lying on the boundary of $\Psi^m$ and the DM is not interested in the boundary, we use \pref{eq:boundaryshift} to determine the step-length for shifting this boundary reference point onto the ROI (line 11 of \pref{alg:NUMS}); otherwise we use \pref{eq:delta} and \pref{eq:upsilon} to serve this purpose (line 13 and line 14 of \pref{alg:NUMS}). At the end of this loop, we use \pref{eq:newposition} to calculate the new position of the biased reference point (line 15 of \pref{alg:NUMS}).

%% file: empirical.tex

\section{Proof-of-Principle Results}
\label{sec:proofofprinciple}

In this section, we empirically validate the effectiveness of the NUMS for assisting the decomposition-based EMO algorithms seek the DM's preferred solutions on the problem instances with two to ten objectives. Our recently proposed MOEA/D variant based on stable matching model, named MOEA/D-STM~\cite{MOEAD-STM} is used as the baseline algorithm. Different from the canonical MOEA/D, where the selection of the next parents is merely determined by the aggregation function value of a solution, MOEA/D-STM treats subproblems and solutions as two sets of agents and considers their mutual-preferences simultaneously. In particular, the preference of a subproblem over a solution measures the convergence issue, while the preference of a solution over a subproblem measures the diversity issue. Since the stable matching achieves an equilibrium of the mutual-preferences between subproblems and solutions, MOEA/D-STM strikes a balance between convergence and diversity of the search process. Here we use the simulated binary crossover (SBX)~\cite{SBX} and the polynomial mutation~\cite{PolyMutation} as the reproduction operators. For the SBX, the crossover probability is set as $p_c=1.0$ and its distribution index is set as $\eta_c=10$; for the polynomial mutation, the mutation probability is set as $p_m=\frac{1}{n}$ and its distribution index is set as $\eta_m=20$. ZDT~\cite{ZDT} and DTLZ~\cite{DTLZ} problem suites are chosen to form the benchmark.

Generally speaking, the proof-of-principle studies consist of two parts. First of all, we validate the effectiveness of the NUMS on the problem instances with two and three objectives. Afterwards, we empirically demonstrate some interesting extensions of the NUMS for handling various other scenarios, i.e., problems with a large number of objectives, multiple ROIs and an interactive preference incorporation.

\subsection{Problems with Two and Three Objectives}
\label{sec:lowdimension}

\begin{figure*}[htbp]
\centering

\pgfplotsset{every axis/.append style={
grid = major,
xlabel = {$f_1$},
ylabel = {$f_2$},
xmin = 0, xmax = 1.001,
ymin = 0, ymax = 1.005,
thick,
line width = 1pt,
tick style = {line width = 0.8pt}}}

  \subfloat[$\tau=0.1$]{
  \resizebox{0.25\textwidth}{!}{
      \begin{tikzpicture}
	\begin{axis}[
      legend columns = -1,
      legend style   = {font=\scriptsize},
      legend entries = {PF, MOEA/D-STM, reference points, $\mathbf{z}^r$= {(0.3, 0.4)}},
      legend to name = named
      ]
		\addplot[very thick, gray, mark = none] table {Data/PF/ZDT1.dat};
    \addplot[only marks, thick, black, mark = *] table {Data/ZDT4/t1.dat};
    \addplot[only marks, thick, gray, mark = o] table {Data/refp/pW2D_ZDT4_1.dat};
    \addplot[only marks, thick, red, mark size = 3.0, mark = star] coordinates {(0.3, 0.4)};
	\end{axis}
      \end{tikzpicture}
    }}
  \subfloat[$\tau=0.3$]{
  \resizebox{0.25\textwidth}{!}{
      \begin{tikzpicture}
	\begin{axis}
    \addplot[very thick, gray, mark = none] table {Data/PF/ZDT1.dat};
    \addplot[only marks, thick, black, mark = *] table {Data/ZDT4/t3.dat};
    \addplot[only marks, thick, gray, mark = o] table {Data/refp/pW2D_ZDT4_3.dat};
    \addplot[only marks, thick, red, mark size = 3.0, mark = star] coordinates {(0.3, 0.4)};
	\end{axis}
      \end{tikzpicture}
    }}
  \subfloat[$\tau=0.7$]{
  \resizebox{0.25\textwidth}{!}{
      \begin{tikzpicture}
	\begin{axis}
    \addplot[very thick, gray, mark = none] table {Data/PF/ZDT1.dat};
    \addplot[only marks, thick, black, mark = *] table {Data/ZDT4/t7.dat};
    \addplot[only marks, thick, gray, mark = o] table {Data/refp/pW2D_ZDT4_7.dat};
    \addplot[only marks, thick, red, mark size = 3.0, mark = star] coordinates {(0.3, 0.4)};
	\end{axis}
      \end{tikzpicture}
    }}
  \subfloat[$\tau=0.9798$]{
  \resizebox{0.25\textwidth}{!}{
      \begin{tikzpicture}
	\begin{axis}
    \addplot[very thick, gray, mark = none] table {Data/PF/ZDT1.dat};
    \addplot[only marks, thick, black, mark = *] table {Data/ZDT4/t10.dat};
    \addplot[only marks, thick, gray, mark = o] table {Data/refp/pW2D_ZDT4_10.dat};
    \addplot[only marks, thick, red, mark size = 3.0, mark = star] coordinates {(0.3, 0.4)};
	\end{axis}
      \end{tikzpicture}
    }}

\ref{named}
\caption{Comparisons of solutions obtained by MOEA/D-STM with different $\tau$ settings for NUMS on ZDT1 problem.}
\label{fig:proof_ZDT1}
\end{figure*}
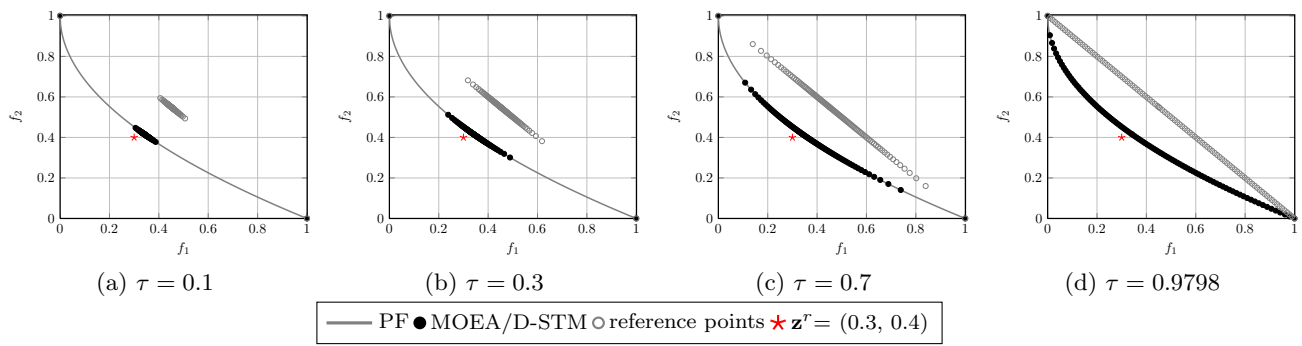

Let us start from the two-objective ZDT1 problem instance that has a convex PF~\cite{ZDT}. The population size of MOEA/D-STM is set to 100 and it performs 300 generations. \pref{fig:proof_ZDT1} shows a comparative results of solutions obtained by MOEA/D-STM with different $\tau$ settings. From this figure, we clearly see that the NUMS adapts the originally uniformly distributed reference points to a biased distribution according to the required extent. In the meanwhile, MOEA/D-STM provides a well approximation of the partial PF with respect to those biased reference points. Note that, in order to approximate the whole PF without preferences on any particular region, we need to set $\tau=1-\frac{2}{99}$ rather than 1.0 according to \pref{theorem:eta}.

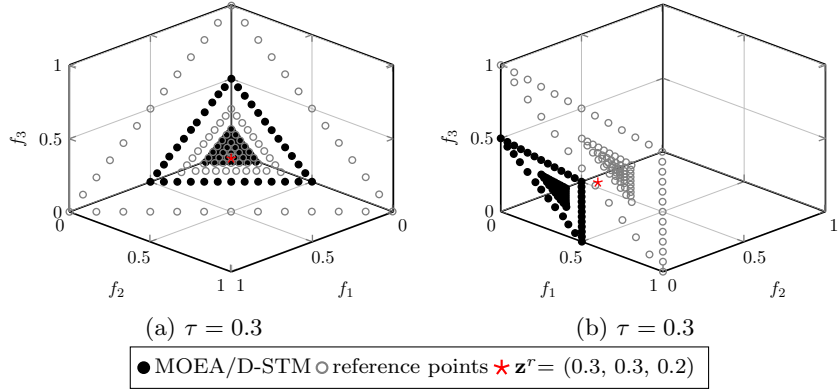
\begin{figure}[htbp]
\centering

\pgfplotsset{every axis/.append style={
grid = major,
view   = {135}{30},
xlabel = {$f_1$},
ylabel = {$f_2$},
zlabel = {$f_3$},
xmin = 0, xmax = 1.001,
ymin = 0, ymax = 1.005,
zmin = 0, zmax = 1.005,
thick,
line width = 1pt,
tick style = {line width = 0.8pt}}}

  \subfloat[$\tau=0.3$]{
  \resizebox{0.33\textwidth}{!}{
      \begin{tikzpicture}
	\begin{axis}[
      legend columns = -1,
      legend style   = {font=\scriptsize},
      legend entries = {MOEA/D-STM, reference points, $\mathbf{z}^r$= {(0.3, 0.3, 0.2)}},
      legend to name = named1
      ]
    \addplot3[only marks, thick, black, mark = *] table {Data/DTLZ1/t3.dat};
    \addplot3[only marks, thick, gray, mark = o] table {Data/refp/pW3D_DTLZ1_3.dat};
    \addplot3[only marks, thick, red, mark size = 3.0, mark = star] coordinates {(0.3, 0.3, 0.2)};
	\end{axis}
      \end{tikzpicture}
    }}
  \subfloat[$\tau=0.3$]{
  \resizebox{0.33\textwidth}{!}{
      \begin{tikzpicture}
	\begin{axis}[
    view = {45}{30}
    ]
    \addplot3[only marks, thick, black, mark = *] table {Data/DTLZ1/t3.dat};
    \addplot3[only marks, thick, gray, mark = o] table {Data/refp/pW3D_DTLZ1_3.dat};
    \addplot3[only marks, thick, red, mark size = 3.0, mark = star] coordinates {(0.3, 0.3, 0.2)};
	\end{axis}
      \end{tikzpicture}
    }}

\ref{named1}
\caption{Solutions obtained by MOEA/D-STM on DTLZ1 problem where $\mathbf{z}^r=(0.3, 0.3, 0.2)^T$.}
\label{fig:proof_DTLZ1}
\end{figure}

\begin{figure}[htbp]
\centering

\pgfplotsset{every axis/.append style={
grid = major,
view   = {135}{30},
xlabel = {$f_1$},
ylabel = {$f_2$},
zlabel = {$f_3$},
xmin = 0, xmax = 1.001,
ymin = 0, ymax = 1.005,
zmin = 0, zmax = 1.005,
thick,
line width = 1pt,
tick style = {line width = 0.8pt}}}

  \subfloat[$\tau=0.3$]{
  \resizebox{0.33\textwidth}{!}{
      \begin{tikzpicture}
	\begin{axis}[
      legend columns = -1,
      legend style   = {font=\scriptsize},
      legend entries = {MOEA/D-STM, reference points, $\mathbf{z}^r$= {(0.3, 0.5, 0.6)}},
      legend to name = named2
      ]
    \addplot3[only marks, thick, black, mark = *] table {Data/DTLZ2/t3.dat};
    \addplot3[only marks, thick, gray, mark = o] table {Data/refp/pW3D_DTLZ2_3.dat};
    \addplot3[only marks, thick, red, mark size = 3.0, mark = star] coordinates {(0.2, 0.5, 0.6)};
	\end{axis}
      \end{tikzpicture}
    }}
  \subfloat[$\tau=0.3$]{
  \resizebox{0.33\textwidth}{!}{
      \begin{tikzpicture}
	\begin{axis}[
    view = {75}{30}
    ]
    \addplot3[only marks, thick, black, mark = *] table {Data/DTLZ2/t3.dat};
    \addplot3[only marks, thick, gray, mark = o] table {Data/refp/pW3D_DTLZ2_3.dat};
    \addplot3[only marks, thick, red, mark size = 3.0, mark = star] coordinates {(0.2, 0.5, 0.6)};
	\end{axis}
      \end{tikzpicture}
    }}

\ref{named2}
\caption{Solutions obtained by MOEA/D-STM on DTLZ2 problem where $\mathbf{z}^r=(0.2, 0.5, 0.6)^T$.}
\label{fig:proof_DTLZ2}
\end{figure}
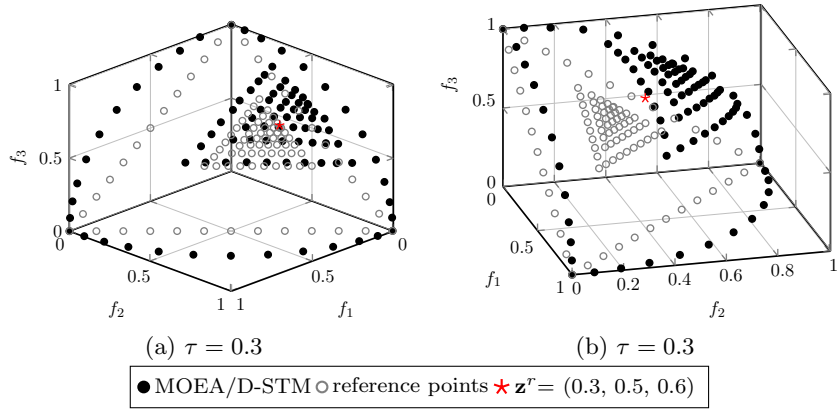

Next, we assess the performance of MOEA/D-STM with the NUMS on the three-objective DTLZ1 and DTLZ2 problem instances respectively. Here we set $\tau=0.2$ in the NUMS, and MOEA/D-STM performs 300 generations for DTLZ2 and 1,000 generations for DTLZ1 due to its multimodality. As shown in~\pref{fig:proof_DTLZ1} and \pref{fig:proof_DTLZ2}, with either an infeasible ($\mathbf{z}^r=(0.3,0.3,0.2)^T$ for DTLZ1) or feasible ($\mathbf{z}^r=(0.3,0.5,0.6)^T$ for DTLZ2) aspiration level vector, MOEA/D-STM has no difficulty in finding the preferred solutions in the ROI. Furthermore, MOEA/D-STM also well approximates the boundaries for both cases.

\begin{figure*}
\centering
\subfloat[MOEA/D-STM.]{\includegraphics[width=.33\linewidth]{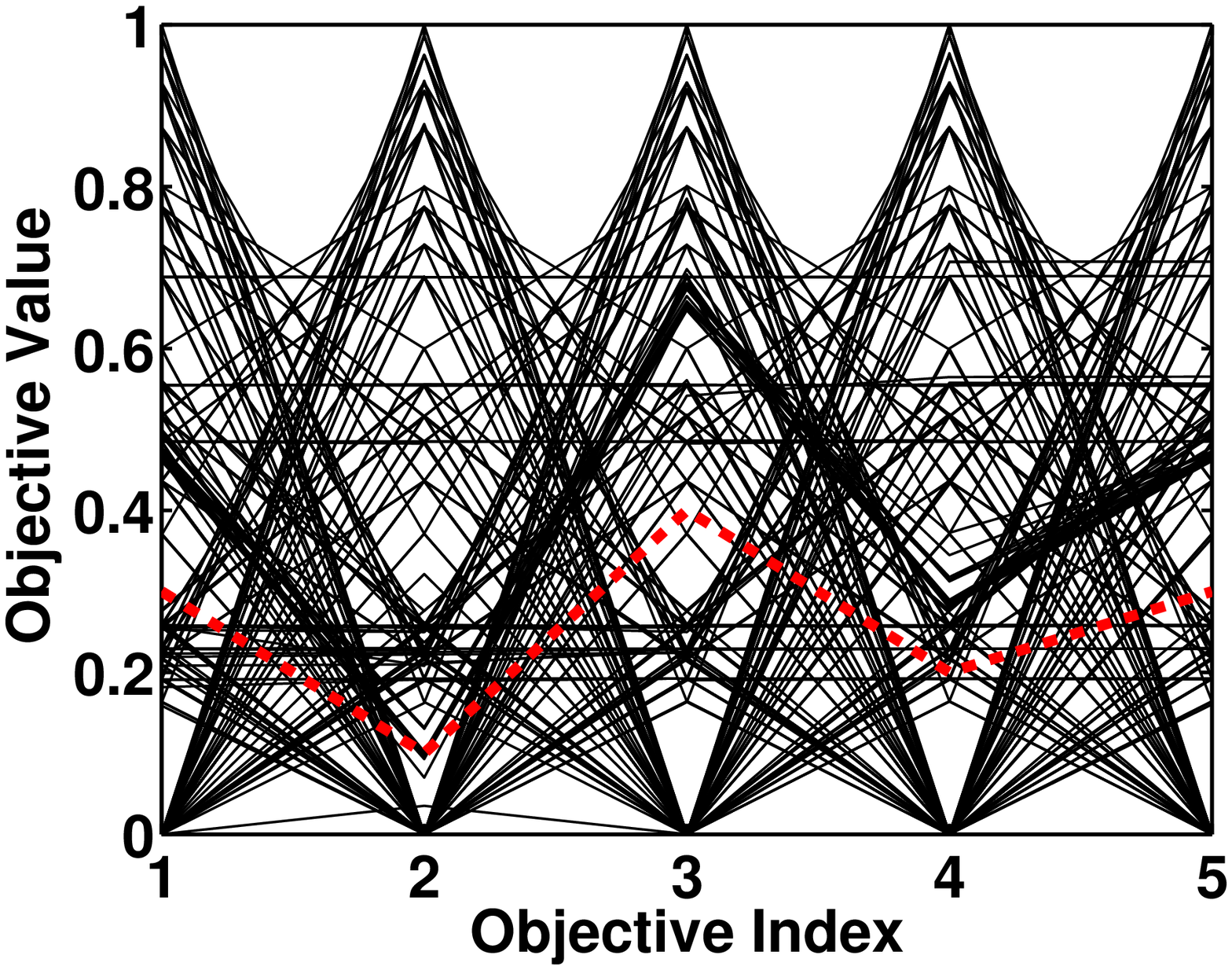}}
\subfloat[Reference points.]{\includegraphics[width=.33\linewidth]{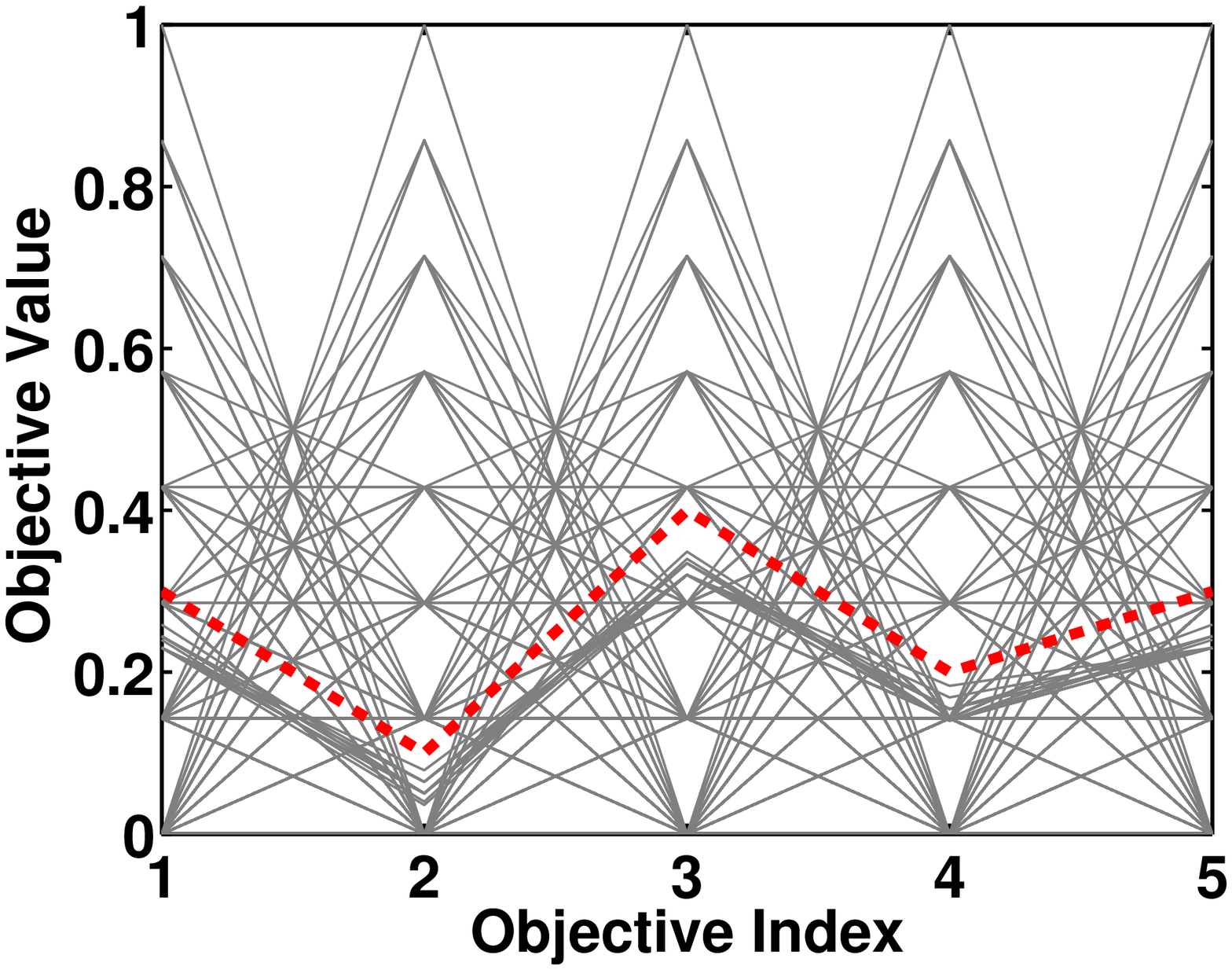}}
\subfloat[Theoretical optima.]{\includegraphics[width=.33\linewidth]{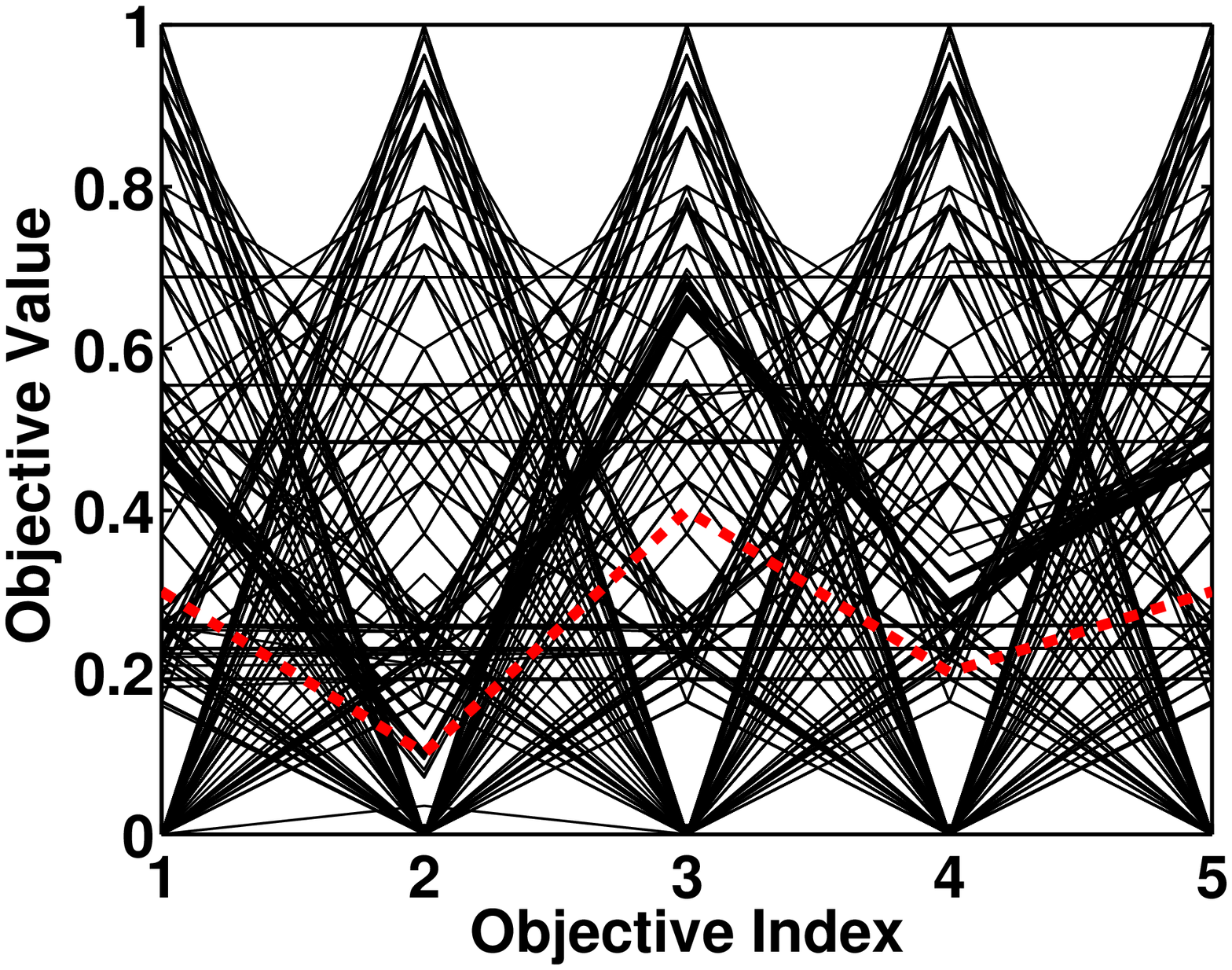}}
\caption{Solutions obtained by MOEA/D-STM on 5-objective DTLZ2 problem where $\mathbf{z}^r=(0.3, 0.1, 0.4, 0.2, 0.3)^T$ is represented as the red dotted line.}
\label{fig:proof_5D}
\end{figure*}

\subsection{Problems with a Large Number of Objectives}
\label{sec:manyobjs}

Now let us consider the five-objective DTLZ2 problem instance where the aspiration level vector is set as $\mathbf{z}^r=(0.3,0.1,0.4,0.2,0.3)^T$. Now, we set $H=7$ in the Das and Dennis's method which generates 330 uniformly distributed reference points, and $\tau$ is set to 0.1 in the NUMS. \pref{fig:proof_5D}(c) gives the theoretical optimal points, with respect to the adapted reference points given in~\pref{fig:proof_5D}(b), on the PF of DTLZ2, according to the method developed in~\cite{MOEADD}. Comparing \pref{fig:proof_5D}(a) with \pref{fig:proof_5D}(c), we can see that MOEA/D-STM, after performing 1,000 generations, has a well approximation to both the ROI and the boundary points.

\begin{figure*}
\centering
\subfloat[MOEA/D-STM.]{\includegraphics[width=.33\linewidth]{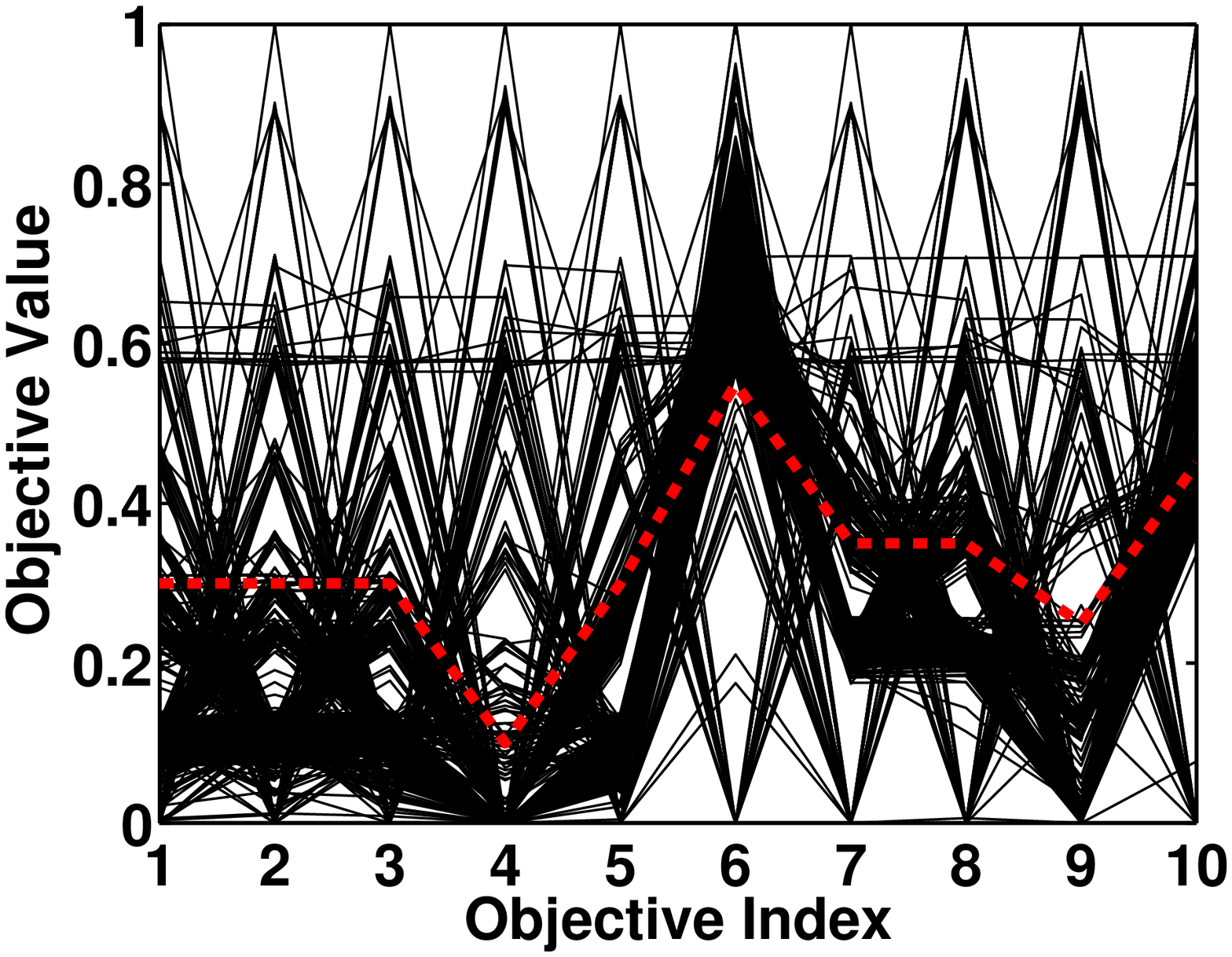}}
\subfloat[Reference points.]{\includegraphics[width=.33\linewidth]{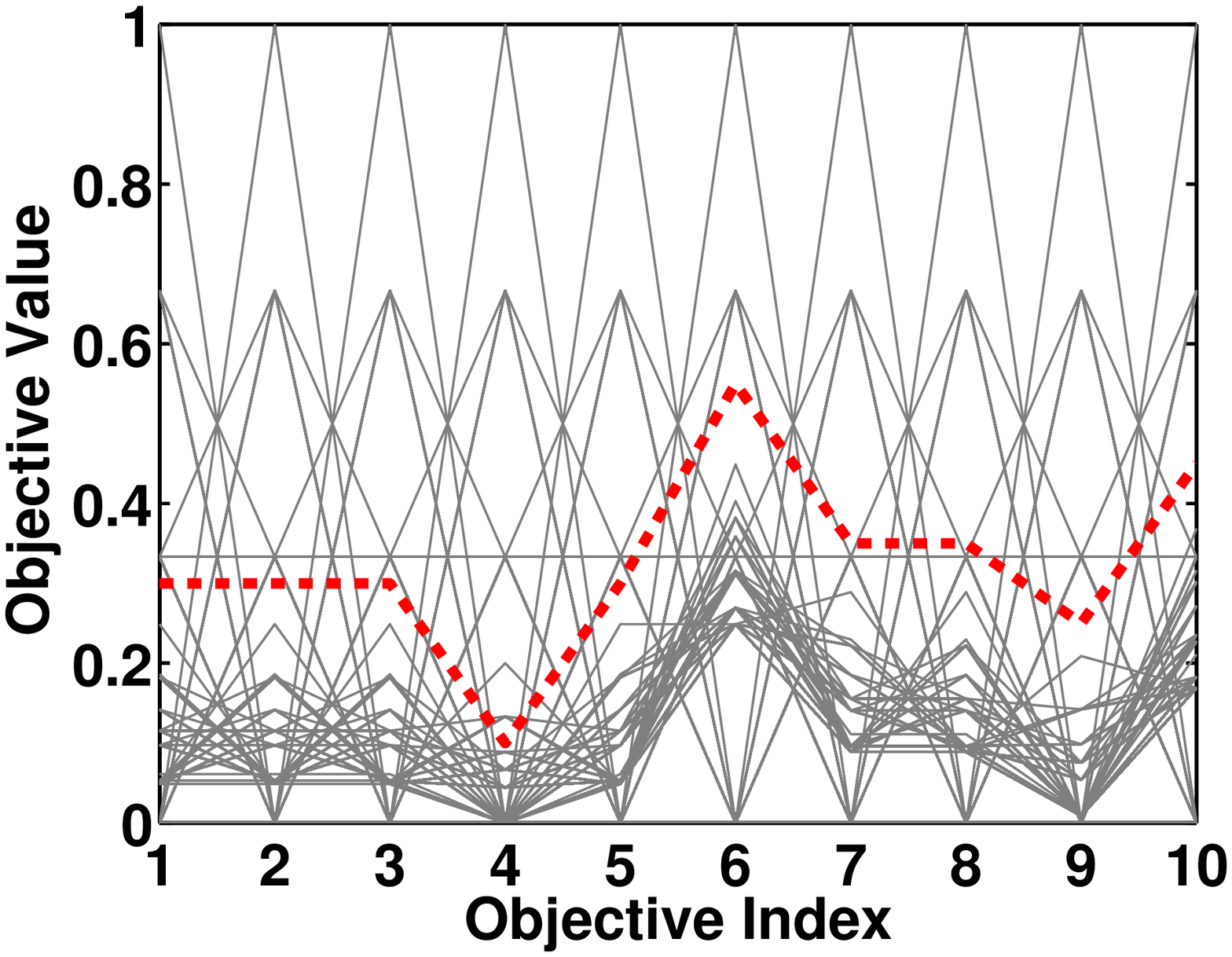}}
\subfloat[Theoretical optima.]{\includegraphics[width=.33\linewidth]{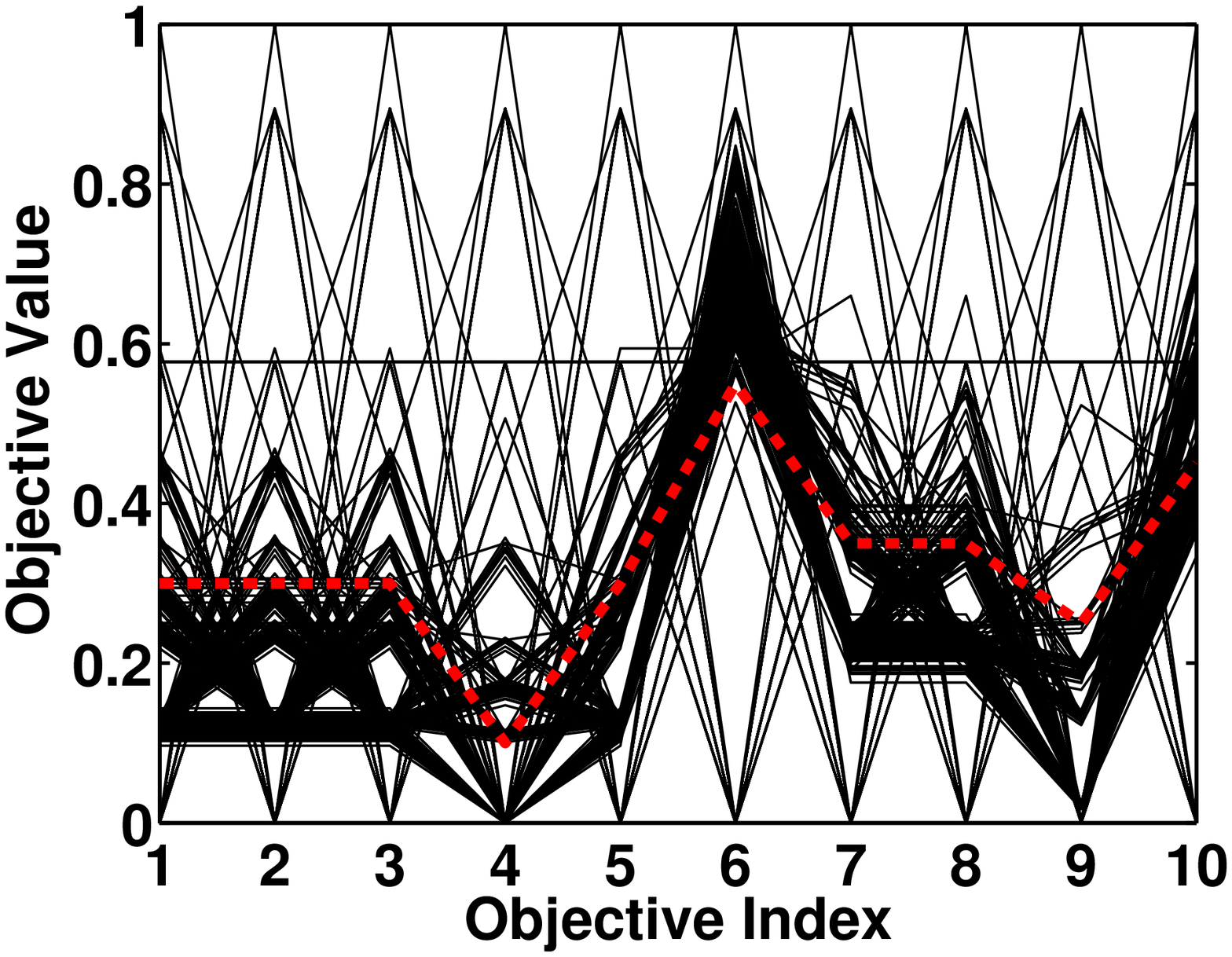}}
\caption{Solutions obtained by MOEA/D-STM on 10-objective DTLZ2 problem where $\mathbf{z}^r=(0.3, 0.3, 0.3, 0.1, 0.3, 0.55, 0.35, 0.35, 0.25, 0.45)^T$ is represented as the red dotted line.}
\label{fig:proof_10D}
\end{figure*}

As discussed in~\cite{MOEADD} and~\cite{NSGA-III}, in order to have intermediate reference points within the simplex, we should set $H\geq m$ in the Das and Dennis's method. Otherwise, all reference points should lie on the boundary of the simplex. However, in a large-dimensional space, we can have a huge amount of reference points even when $H=m$. For example, when $m=10$, the Das and Dennis's method can generate $\binom{10+10-1}{10-1}=92378$ uniformly distributed reference points if $H=10$. Obviously, current EMO algorithms cannot hold such huge number of solutions in a population. Even worse, when $H=m$, there is only one intermediate reference point which lies in the center of the simplex. Thus, the original NUMS might not be directly applicable when facing a large number of objectives. Inspired by the multi-layer weight vector generation method developed in~\cite{MOEADD} and~\cite{NSGA-III}, we make a slight modification to adapt the NUMS to the many-objective scenario. First of all, we use the Das and Dennis's method, where $H<m$, more than one time, say $l>1$, to generate $l$ layers of reference points. Afterwards, we use the method developed in~\pref{sec:boundary} to shift these reference points, which lie on the boundary of the simplex, onto the ROI layer by layer. \pref{fig:proof_10D}(b) shows an example of 661 reference points generated by the multi-layer NUMS where the aspiration level vector is set as $\mathbf{z}^r=(0.3, 0.3, 0.3, 0.1, 0.3, 0.55, 0.35, 0.35, 0.25, 0.45)^T$. In particular, we first use the Das and Dennis's method to generate $l=3$ layers of reference points. Since we set $H=3$, each layer contains 220 reference points. Then two layers of them are shifted onto the ROI, where the shrinkage factor $\tau$ is set to 0.4 and 0.2 respectively. \pref{fig:proof_10D}(a) shows the final solutions obtained by MOEA/D-STM after 1,000 generations. Comparing to the theoretical optima shown in~\pref{fig:proof_10D}(c), we can see that MOEA/D-STM still has a satisfactory approximation to the ROI in a 10-objective space.

\subsection{Investigations on multiple ROIs}
\label{sec:multipleROIs}

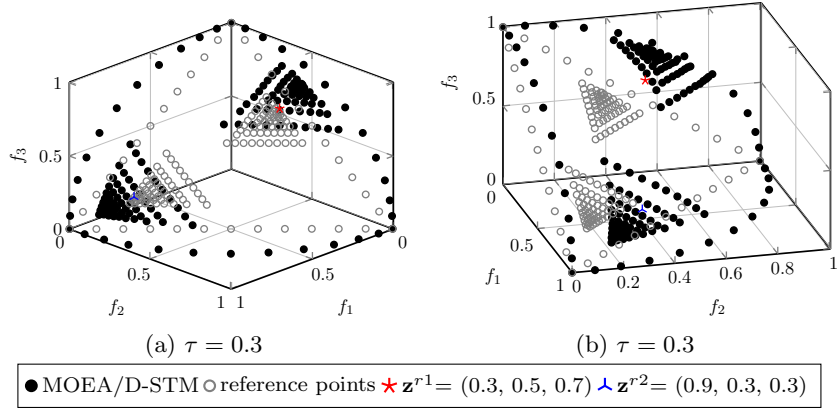
\begin{figure}[htbp]
\centering

\pgfplotsset{every axis/.append style={
grid = major,
view   = {135}{30},
xlabel = {$f_1$},
ylabel = {$f_2$},
zlabel = {$f_3$},
xmin = 0, xmax = 1.001,
ymin = 0, ymax = 1.005,
zmin = 0, zmax = 1.005,
thick,
line width = 1pt,
tick style = {line width = 0.8pt}}}

  \subfloat[$\tau=0.3$]{
  \resizebox{0.33\textwidth}{!}{
      \begin{tikzpicture}
	\begin{axis}[
      legend columns = -1,
      legend style   = {font=\scriptsize},
      legend entries = {MOEA/D-STM, reference points, $\mathbf{z}^{r1}$= {(0.3, 0.5, 0.7)}, $\mathbf{z}^{r2}$= {(0.9, 0.3, 0.3)}},
      legend to name = named3
      ]
    \addplot3[only marks, thick, black, mark = *] table {Data/DTLZ2/t2_double.dat};
    \addplot3[only marks, thick, gray, mark = o] table {Data/refp/pW3D_DTLZ2_double.dat};
    \addplot3[only marks, thick, red, mark size = 3.0, mark = star] coordinates {(0.2, 0.5, 0.7)};
    \addplot3[only marks, thick, blue, mark size = 3.0, mark = Mercedes star] coordinates {(0.9, 0.3, 0.3)};
	\end{axis}
      \end{tikzpicture}
    }}
  \subfloat[$\tau=0.3$]{
  \resizebox{0.33\textwidth}{!}{
      \begin{tikzpicture}
	\begin{axis}[
    view = {75}{30}
    ]
    \addplot3[only marks, thick, black, mark = *] table {Data/DTLZ2/t2_double.dat};
    \addplot3[only marks, thick, gray, mark = o] table {Data/refp/pW3D_DTLZ2_double.dat};
    \addplot3[only marks, thick, red, mark size = 3.0, mark = star] coordinates {(0.2, 0.5, 0.7)};
    \addplot3[only marks, thick, blue, mark size = 3.0, mark = Mercedes star] coordinates {(0.9, 0.3, 0.3)};
	\end{axis}
      \end{tikzpicture}
    }}

\ref{named3}
\caption{Solutions obtained by MOEA/D-STM with two different reference points on DTLZ2 problem.}
\label{fig:proof_DTLZ2double}
\end{figure}

In practice, the DM might not be sure about his/her exact preferences and he/she would like to simultaneously explore several ROIs. In this case, the DM would like to supply more than one, say $T>1$, aspiration level vectors simultaneously. To accommodate multiple ROIs, we only need to apply the NUMS $T$ times with respect to each aspiration level vector. Note that each time the NUMS can preserve the boundary reference points, but we only need to keep these boundary reference points once. In other words, the duplicated boundary reference points are exempted from the final reference point set. \pref{fig:proof_DTLZ2double} shows an example of two aspiration level vectors. In particular, the gray points are the adapted reference points for each ROI; while the black points are the final solutions obtained by MOEA/D-STM with respect to the corresponding reference points. From the experimental results, we can clearly see that MOEA/D-STM with the NUMS is also able to approximate multiple ROIs simultaneously.

\subsection{Interactive Scenario}
\label{sec:interactive}

\begin{figure*}[htbp]
\centering

\pgfplotsset{every axis/.append style={
grid = major,
view   = {75}{30},
xlabel = {$f_1$},
ylabel = {$f_2$},
zlabel = {$f_3$},
xmin = 0, xmax = 1.001,
ymin = 0, ymax = 1.005,
zmin = 0, zmax = 1.005,
thick,
line width = 1pt,
tick style = {line width = 0.8pt}}}

\subfloat[1st interaction]{
  \resizebox{0.33\textwidth}{!}{
\begin{tikzpicture}
\begin{axis}[
  legend columns = -1,
  legend style   = {font=\scriptsize},
  legend entries = {MOEA/D-STM, reference points},
  legend to name = named4,
  xmax = 2.000,
  ymax = 2.000,
  zmax = 2.000
  ]
  \addplot3[only marks, thick, black, mark = *] table {Data/IN/FUN0.dat};
  \addplot3[only marks, thick, gray, mark = o] table {Data/refp/pW3D_DTLZ2_IN_1.dat};
  \addplot3[only marks, thick, red, mark size = 3.0, mark = star] coordinates {(1.4, 1.9, 1.5)};
\end{axis}
\end{tikzpicture}
}}
\subfloat[2nd interaction]{
  \resizebox{0.33\textwidth}{!}{
\begin{tikzpicture}
\begin{axis}
  \addplot3[only marks, thick, black, mark = *] table {Data/IN/FUN1.dat};
  \addplot3[only marks, thick, gray, mark = o] table {Data/refp/pW3D_DTLZ2_IN_2.dat};
  \addplot3[only marks, thick, red, mark size = 3.0, mark = star] coordinates {(0.7, 0.6, 0.3)};
\end{axis}
\end{tikzpicture}
}}
\subfloat[3rd interaction]{
  \resizebox{0.33\textwidth}{!}{
\begin{tikzpicture}
\begin{axis}
  \addplot3[only marks, thick, black, mark = *] table {Data/IN/FUN2.dat};
  \addplot3[only marks, thick, gray, mark = o] table {Data/refp/pW3D_DTLZ2_IN_3.dat};
  \addplot3[only marks, thick, red, mark size = 3.0, mark = star] coordinates {(0.3, 0.4, 0.8)};
\end{axis}
\end{tikzpicture}
}}

\ref{named4}
\caption{Interactive scenario on DTLZ2 problem.}
\label{fig:interactive}
\end{figure*}
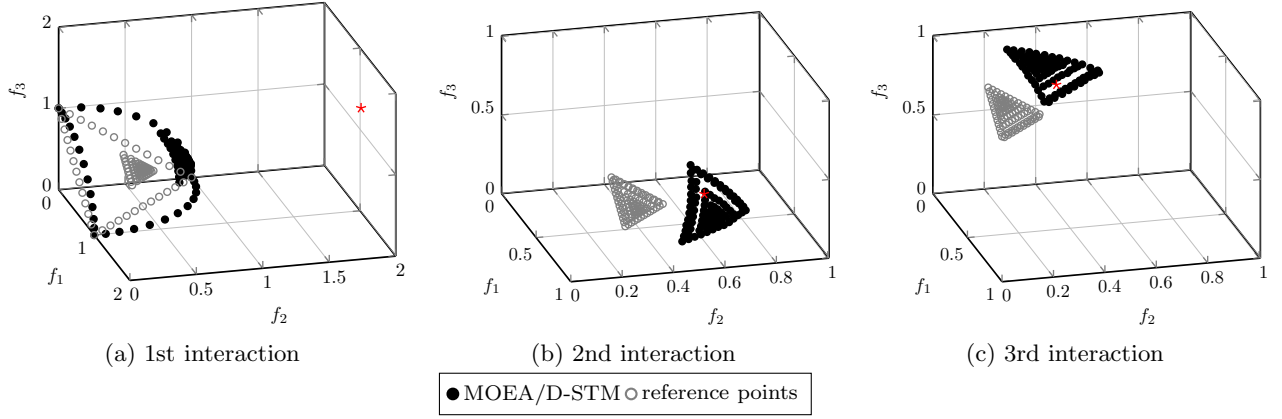

In real-world optimization, the DM is usually not fully confident about his/her elicited preference information due to the lack of the knowledge of the PF beforehand. Therefore, an interactive decision making procedure where the DM can progressively adapt his/her preference information during the optimization process is attractive in the preference-based EMO. Since the NUMS can easily adapt the distribution of reference points to be biased towards the ROI, it facilitates the interactive scenario. Moreover, also due to the lack of the knowledge of the PF, the DM can easily specify an aspiration level vector which is far beyond the boundary of the PF. Since the NUMS is able to preserve the boundary reference points, it finally helps the DM better understand the PF (e.g., its general shape, boundary, ideal and nadir points) and further adjusts his/her preference information. In~\pref{fig:interactive}, we describe an interactive run, which includes three cycles, of the MOEA/D-STM on the DTLZ2 problem. We call a run of the MOEA/D-STM for a certain number of generations specified by the DM as a cycle. First, as shown in~\pref{fig:interactive}(a), the DM specifies an aspiration level vector $\mathbf{z}^r_1=(1.4,1.9,1.5)^T$ far beyond the PF. After 200 generations, MOEA/D-STM finds the solutions not only crowd in the ROI, but also distribute along the boundary. Thereafter, the DM realizes that $\mathbf{z}^r_1$ is a bad choice, and then he/she resets another aspiration level vector, say $\mathbf{z}^r_2=(0.7,0.6,0.3)^T$. In addition, since the DM already knows the boundary of the PF, he/she might not be interested in the boundary any longer. Thus, he/she sets the NUMS to adapt all reference points to the ROI. By using the final population of the first interaction as the initial population, MOEA/D-STM finally finds the solutions in the ROI after 200 generations. However, we assume that the DM still does not satisfy them and he/she sets another aspiration level vector, say $\mathbf{z}^r_3=(0.3, 0.4, 0.8)^T$. After 200 generations, as shown in~\pref{fig:interactive}(c), MOEA/D-STM finds the solutions in the vicinity of the ROI. This time, the DM is comfortable with the obtained solutions and the interactive EMO process terminates.

\section{Comparisons with the State-of-the-Art}
\label{sec:comparisons}

The proof-of-principle results shown in~\pref{sec:proofofprinciple} fully demonstrate the effectiveness of the NUMS for assisting a decomposition-based EMO algorithm (here we use MOEA/D-STM as an example) search for preferred solutions in the ROI. As discussed in~\pref{sec:relatedworks}, there are some other preference-based EMO algorithms proposed in the EMO literature. In this section, we compare the performance of MOEA/D-STM assisted by the NUMS, where $\tau$ is set as 0.2, with three state-of-the-art preference-based EMO algorithms, i.e., g-NSGA-II~\cite{MolinaSDCC09}, R-NSGA-II~\cite{DebSBC06} and r-NSGA-II~\cite{SaidBG10}. Note that all these preference-based EMO algorithms use the aspiration level vector to elucidate the DM's preference information. In addition, it might also be interesting to see how does the performance of a general purpose EMO algorithm, which is developed to approximate the whole PF, compare with a preference-based EMO algorithm. To this end, we consider using NSGA-III~\cite{DebJ14}, a state-of-the-art many-objective optimizer, as the representative of the general purpose EMO algorithm in our experiments. All the multi-objective optimizers use the SBX and the polynomial mutation for offspring generation. The corresponding parameters are set the same as~\pref{sec:proofofprinciple}. In the experiments, we choose the popular DTLZ1 to DTLZ4 as the test problems, where $m\in\{3,5,8,10\}$. \pref{tab:settings-reference} provides the settings of two kinds of aspiration level vectors, one is unattainable while the other is attainable. The population size is set as $N=92$ when $m=3$; $N=210$ when $m=5$; $N=360$ when $m=8$; and $N=660$ when $m=10$, respectively. As for the NUMS, the number of divisions is set as $H=13$ when $m=3$ and $H=6$ when $m=5$. When the number of objectives is larger than 5, we use a 3-layer method suggested in~\pref{sec:manyobjs} to generate the initially uniformly distributed reference points. In particular, we set $H=3$ for each layer. In our experiments, the stopping criterion of a preference-based EMO algorithm is the number of function evaluations (FEs), as presented in~\pref{tab:settings-generations}. As for R-NSGA-II, the additional parameter $\epsilon$, used in its $\epsilon$-clearing procedure, is set according to~\cite{DebSBC06}, i.e., $\epsilon=0.01$.

\begin{table*}[htbp]
\scriptsize
\centering
\caption{Settings of Aspiration Level Vector}
\label{tab:settings-reference}
\begin{tabular}{c|c|c|c}
\hline
Test Problem & $m$              & Unattainable                                                                 & Attainable                                                 \\\hline
                                & 3         & $(0.05, 0.05, 0.2)^T$                                            & $(0.3, 0.3, 0.2)^T$                                        \\\cline{2-4}
\multirow{2}{*}{DTLZ1}          & 5         & $(0.05, 0.05, 0.1, 0.08, 0.03)^T$                                & $(0.2, 0.1, 0.1, 0.3, 0.4)^T$                              \\\cline{2-4}
                                & 8         & $(0.01, 0.02, 0.07, 0.02, 0.06, 0.2, 0.1, 0.01)^T$               & $(0.1, 0.2, 0.1, 0.4, 0.4, 0.1, 0.3, 0.1)^T$               \\\cline{2-4}
                                & 10        & $(0.02, 0.01, 0.06, 0.04, 0.04, 0.01, 0.02, 0.03, 0.05, 0.08)^T$ & $(0.05, 0.1, 0.1, 0.05, 0.1, 0.2, 0.08, 0.03, 0.3, 0.1)^T$ \\\hline
                                & 3         & $(0.2, 0.5, 0.6)^T$                                              & $(0.7, 0.8, 0.5)^T$                                        \\\cline{2-4}
\multirow{2}{*}{DTLZ2 to DTLZ4} & 5         & $(0.3, 0.1, 0.4, 0.2, 0.3)^T$                                    & $(0.7, 0.6, 0.3, 0.8, 0.5)^T$                              \\\cline{2-4}
                                & 8         & $(0.3, 0.1, 0.4, 0.25, 0.1, 0.15, 0.4, 0.25)^T$                  & $(0.6, 0.5, 0.75, 0.2, 0.3, 0.55, 0.7, 0.6)^T$             \\\cline{2-4}
                                & 10        & $(0.1, 0.1, 0.3, 0.4, 0.2, 0.5, 0.25, 0.15, 0.1, 0.4)^T$         & $(0.3, 0.3, 0.3, 0.1, 0.3, 0.55, 0.35, 0.35, 0.25, 0.45)^T$   \\\hline
\end{tabular}
\end{table*}

\begin{table}[htbp]
\scriptsize
\centering
\caption{Settings of Number of Function Evaluations}
\label{tab:settings-generations}
\begin{tabular}{c|c|c|c|c|c}
\hline
Test Problem           & $m$ & \# of FEs      & Test Problem           & $m$ & \# of FEs      \\\hline
                       & 3   & $400\times N$  &                        & 3   & $1000\times N$ \\\cline{2-3}\cline{5-6}
\multirow{2}{*}{DTLZ1} & 5   & $1000\times N$ & \multirow{2}{*}{DTLZ3} & 5   & $1200\times N$ \\\cline{2-3}\cline{5-6}
                       & 8   & $1200\times N$ &                        & 8   & $1500\times N$ \\\cline{2-3}\cline{5-6}
                       & 10  & $1500\times N$ &                        & 10  & $1800\times N$ \\\hline
                       & 3   & $250\times N$  &                        & 3   & $600\times N$  \\\cline{2-3}\cline{5-6}
\multirow{2}{*}{DTLZ2} & 5   & $800\times N$  & \multirow{2}{*}{DTLZ4} & 5   & $1200\times N$ \\\cline{2-3}\cline{5-6}
                       & 8   & $1000\times N$ &                        & 8   & $1500\times N$ \\\cline{2-3}\cline{5-6}
                       & 10  & $1360\times N$ &                        & 10  & $1800\times N$ \\\hline
\end{tabular}
\end{table}

For a quantitative comparison, we use our recently developed R-metrics~\cite{LiD16} to evaluate the quality of an approximation set. The basic idea of R-metric evaluation is to pre-process the approximation sets found by different algorithms before using the inverted generational distance (IGD)~\cite{IGD} and hypervolume (HV)~\cite{SPEA} for performance assessment. Here we only give a general idea of the R-metric calculation while more interested readers can refer to~\cite{LiD16} for details.
\begin{itemize}
    \item\textbf{Step 1. Representative Point Identification:} We use the solution closest to the centroid of the underlying solution set as the representative point $\overline{\mathbf{z}}$.
    \item\textbf{Step 2. Filtering:} We only keep the solutions, which are close to $\overline{\mathbf{z}}$ and within a relative extent of the ROI, for the R-metric calculation.
    \item\textbf{Step 3. Solution Transfer:} Along the iso-ASF line, we transfer the filtered solutions towards the reference line constructed by the aspiration level vector $\mathbf{z}^r$ and a given worst point $\mathbf{z}^w$. Here we set $\mathbf{z}^w=\mathbf{z}^r+2\times\mathbf{I}$, where $\mathbf{I}$ represents the unit vector, as in~\cite{LiD16}.
    \item\textbf{Step 4. R-metric Calculation:} Calculate the IGD or HV value of the processed solution set as the R-IGD or R-HV value of the underlying approximation set.
\end{itemize}

Similar to the original IGD and HV metrics, the lower is the R-IGD value (or the larger is the R-HV value), the better is the quality of a solution set for approximating the ROI. For the R-metric calculation, we set the relative extent of the ROI as 0.2. As for the R-IGD calculation, 10,000 points are sampled from the corresponding PF when $m=3$; 17,550 points are sampled when $m=5$; 77,520 points are sampled when $m=8$; and 293,930 points are sampled when $m=10$. In the experiments, each algorithm is performed 31 independent runs. In the data tables, we show the mean and variance of R-metric values for different problem instances with various aspiration level vector settings. In particular, the best median metric values are highlighted in bold face with gray background. To have a statistically sound conclusion, we employ the Wilcoxon's rank sum test at a 5\% significance level to validate the statistical significance of the best mean metric values. To have a visual comparison, we also show the parallel coordinate plots (PCP) of the final solutions obtained by different algorithms having the best R-IGD value. Due to the page limit, all these plots are put in the supplementary file.

\subsection{Experimental Results}
\label{sec:results}

From the results shown in~\pref{tab:Rmetric-unattainable} to~\pref{tab:Rmetric-attainable}, we can clearly see that MOEA/D-STM assisted by the NUMS is the best candidate for approximating the ROI on various test problems. In particular, it achieves the best R-metric values in all 256 comparisons and all the better results are statistically significant. In the following paragraphs, we explain the results instance by instance.

Let us start from the DTLZ1 problem which has a linear PF shape, i.e., a hyper-plane intersects with each coordinate at 0.5. Note that DTLZ1 also has many local optima in its search space, which obstruct the convergence towards the global PF. In the 3-objective case, all algorithms, except g-NSGA-II, are able to drive solutions to converge towards the PF. As the DM expects the ROI to be 20\% of the whole PF, solutions found by MOEA/D-STM assisted by the NUMS are the best candidates that satisfies the DM's expectation, as shown in Fig. 2 and Fig. 3 of the supplementary file. In contrast, although the solutions found by R-NSGA-II are in the ROI, they crowd in a narrow region. In this case, R-NSGA-II cannot provide as many trade-off alternatives as MOEA/D-STM assisted by the NUMS. Although NSGA-III obtains a set of solutions that well approximate the whole PF, as discussed in~\pref{sec:introduction}, it provides too many irrelevant solutions which might not be interested by the DM and thus dilute the resolution of the ROI. With the increase of the number of objectives, g-NSGA-II and r-NSGA-II have difficulty in driving the solutions towards the PF due to the multi-modal property of DTLZ1. As for R-NSGA-II, solutions are even more focused in the high-dimensional space as shown in the PCP of the supplementary file. Although NSGA-III provides an acceptable approximation to the whole PF, it provides less preferred solutions in the ROI due to the explosion of the objective space. The high-dimensional PF also causes a severe cognitive obstacle for the decision making.

DTLZ2 is a relatively simple test problem, where the objective functions of a Pareto-optimal solution $\mathbf{x}^{\ast}$ satisfies: $\sum_{i=1}^m=f_i^2(\mathbf{x}^{\ast})=1$. All algorithms, except g-NSGA-II, do not have any difficulty in driving the solutions towards the PF. It is interesting to note that although some solutions found by r-NSGA-II do not converge well to the PF, its performance in the high-dimensional space is satisfactory. We also notice that the performance of r-NSGA-II is similar to R-NSGA-II in 5- and 8-objective cases. Similar to the observations in DTLZ1, NSGA-III approximates the whole PF, thus it provides many irrelevant trade-off alternatives outside the ROI.

The PF of DTLZ3 is the same as DTLZ2. But its search space contains many local optima which can make an EMO algorithm get stuck at any local PF before converging to the global PF. Similar to the observations in DTLZ1, g-NSGA-II cannot find any converged solutions in all 3- to 10-objective cases. The performance of MOEA/D-STM assisted with the NUMS is very robust. It is interesting to note that solutions found by r-NSGA-II do not converge well to the ROI in the 3-objective case. This might be caused by the failure of its adaptive parameter control given a limited number of FEs. We also notice that solutions found by r-NSGA-II do not converge to the PF when the number of objectives becomes large.

DTLZ4 also has the identical PF shape as DTLZ2. However, in order to investigate an EMO algorithm's ability to maintain a good distribution of solutions, DTLZ4 introduces a parametric variable mapping to the objective functions of DTLZ2. This modification allows a biased density of points away from $f_m(\mathbf{x})=0$. It is interesting to note that the performance of all these algorithms are similar to the DTLZ2. Specifically, g-NSGA-II cannot drive all solutions converge onto the PF due to the biased density of solutions. In the 3-objective case, some solutions found by r-NSGA-II are still drifted away from the PF when encountering an attainable aspiration level vector. We notice that NSGA-III has some difficulty in optimizing some particular objectives when the number of objectives becomes large. This might be caused by the biased density of solutions which makes the diversity preservation in a high-dimensional space becomes difficult. In this case, it becomes even more risky to use NSGA-III to find preferred solutions.

In summary, the experimental results fully demonstrate the effectiveness of the NUMS for assisting MOEA/D-STM, a representative decomposition-based EMO algorithm, to approximate the ROIs. In the meanwhile, we find that R-NSGA-II is also good at searching for the ROIs, but the spread of the preferred solutions is controlled in an ad-hoc manner. Therefore, R-NSGA-II might not be able to find as many trade-off alternatives as the DMs want. On the other hand, the general purpose EMO algorithm is able to provide some solutions located in the ROIs, especially when the number of objectives is small. However, with the increase of the number of objectives and the problem difficulty, it is risky to use a general purpose EMO algorithm to search for the preferred solutions in the ROIs.

\begin{table*}[htbp]
\tiny
\centering
\caption{Comparison Results of R-IGD and R-HV Values on the Unattainable Aspiration Level Vector}
\label{tab:Rmetric-unattainable}
\begin{tabular}{|c|c|c|c|c|c|c|c|}
\cline{3-7}
\multicolumn{1}{c}{}   &     & \multicolumn{5}{c|}{R-IGD} & \multicolumn{1}{c}{} \\\hline
Test Problem           & $m$ & MOEA/D-STM                                                            & R-NSGA-II         & g-NSGA-II         & r-NSGA-II         & NSGA-III          & $s$      \\\hline
                       & 3   & \multicolumn{1}{>{\columncolor{mycyan}}c}{\textbf{8.947E-2(1.51E-4)}} & 1.329E-1(1.12E-2) & --                & 1.591E-1(6.19E-2) & 9.811E-2(1.73E-2) & ${\dag}$ \\\cline{2-8}
\multirow{2}{*}{DTLZ1} & 5   & \multicolumn{1}{>{\columncolor{mycyan}}c}{\textbf{1.719E-1(8.75E-5)}} & 1.719E-1(1.37E-2) & --                & 2.119E+1(1.34E+2) & 1.758E-1(4.42E-2) & ${\dag}$ \\\cline{2-8}
                       & 8   & \multicolumn{1}{>{\columncolor{mycyan}}c}{\textbf{1.303E-1(5.48E-4)}} & 1.536E-1(6.22E-3) & --                & 6.299E+0(3.56E+0) & 1.991E-1(1.58E-3) & ${\dag}$ \\\cline{2-8}
                       & 10  & \multicolumn{1}{>{\columncolor{mycyan}}c}{\textbf{1.278E-1(8.30E-4)}} & 2.261E-1(2.27E-2) & --                & 2.992E+1(1.10E+1) & 1.622E-1(2.53E-3) & ${\dag}$ \\\hline
                       & 3   & \multicolumn{1}{>{\columncolor{mycyan}}c}{\textbf{8.500E-2(7.94E-4)}} & 1.305E-1(1.55E-2) & 1.305E-1(1.55E-2) & 1.492E-1(1.13E-2) & 4.998E-1(1.07E-2) & ${\dag}$ \\\cline{2-8}
\multirow{2}{*}{DTLZ2} & 5   & \multicolumn{1}{>{\columncolor{mycyan}}c}{\textbf{2.648E-1(1.70E-2)}} & 3.390E-1(4.95E-2) & --                & 4.174E-1(1.14E-1) & 6.123E-1(1.91E-2) & ${\dag}$ \\\cline{2-8}
                       & 8   & \multicolumn{1}{>{\columncolor{mycyan}}c}{\textbf{4.450E-1(1.17E-1)}} & 5.351E-1(8.93E-2) & --                & 5.680E-1(1.56E-1) & 6.215E-1(1.18E-2) & ${\dag}$ \\\cline{2-8}
                       & 10  & \multicolumn{1}{>{\columncolor{mycyan}}c}{\textbf{4.103E-1(7.80E-3)}} & 4.565E-1(4.63E-2) & --                & 5.831E-1(2.15E-1) & 6.553E-1(1.07E-2) & ${\dag}$ \\\hline
                       & 3   & \multicolumn{1}{>{\columncolor{mycyan}}c}{\textbf{8.864E-2(1.05E-2)}} & 1.196E-1(2.07E-2) & 1.414E-1(3.18E-2) & 4.893E-1(8.77E-2) & 5.154E-1(8.47E-3) & ${\dag}$ \\\cline{2-8}
\multirow{2}{*}{DTLZ3} & 5   & \multicolumn{1}{>{\columncolor{mycyan}}c}{\textbf{2.665E-1(1.67E-2)}} & 2.806E-1(5.40E-2) & --                & 2.022E+1(1.06E+1) & 7.043E-1(1.92E-1) & ${\dag}$ \\\cline{2-8}
                       & 8   & \multicolumn{1}{>{\columncolor{mycyan}}c}{\textbf{4.531E-1(5.43E-2)}} & 4.695E-1(6.04E-2) & --                & 4.168E+1(2.05E+1) & 6.324E-1(1.88E-2) & ${\dag}$ \\\cline{2-8}
                       & 10  & \multicolumn{1}{>{\columncolor{mycyan}}c}{\textbf{4.160E-1(1.10E-2)}} & 4.126E-1(6.43E-2) & --                & 9.629E+1(4.47E+1) & 6.538E-1(8.31E-3) & ${\dag}$ \\\hline
                       & 3   & \multicolumn{1}{>{\columncolor{mycyan}}c}{\textbf{8.507E-2(1.10E-3)}} & 1.041E-1(1.64E-2) & 1.374E-1(3.48E-2) & 1.444E-1(7.53E-3) & 6.212E-1(2.22E-1) & ${\dag}$ \\\cline{2-8}
\multirow{2}{*}{DTLZ4} & 5   & \multicolumn{1}{>{\columncolor{mycyan}}c}{\textbf{2.563E-1(1.88E-2)}} & 3.389E-1(5.87E-2) & 2.068E+0(6.41E-1) & 2.950E-1(2.60E-2) & 6.732E-1(2.03E-1) & ${\dag}$ \\\cline{2-8}
                       & 8   & \multicolumn{1}{>{\columncolor{mycyan}}c}{\textbf{4.477E-1(1.59E-2)}} & 5.069E-1(8.94E-2) & 4.690E+0(9.25E-1) & 4.949E-1(4.33E-2) & 6.199E-1(1.17E-2) & ${\dag}$ \\\cline{2-8}
                       & 10  & \multicolumn{1}{>{\columncolor{mycyan}}c}{\textbf{4.165E-1(8.48E-3)}} & 5.123E-1(2.86E-2) & --                & 4.663E-1(9.32E-2) & 6.486E-1(5.13E-3) & ${\dag}$ \\
    \hline
\multicolumn{1}{c}{}   &     & \multicolumn{5}{c|}{R-HV} & \multicolumn{1}{c}{} \\\hline
                       & 3   & \multicolumn{1}{>{\columncolor{mycyan}}c}{\textbf{7.7102(4.99E-1)}}    & 6.9361(1.47E-1)    & --              & 7.2500(2.63E-3)    & 8.0378(8.12E-2)   & ${\dag}$ \\\cline{2-8}
\multirow{2}{*}{DTLZ1} & 5   & 29.4673(2.60E-2)          & 26.6700(7.78E-1)   & --              & 0(0)               & \multicolumn{1}{>{\columncolor{mycyan}}c}{\textbf{29.9567(1.68E+0)}}  &   \\\cline{2-8}
                       & 8   & 249.4514(1.06E+0)  & \multicolumn{1}{>{\columncolor{mycyan}}c}{\textbf{254.3028(4.08E+0)}}  & --              & 3.1855(9.42E+0)    & 214.7997(4.97E+0) & ${\dag}$ \\\cline{2-8}
                       & 10  & \multicolumn{1}{>{\columncolor{mycyan}}c}{\textbf{983.9530(5.09E+0)}}  & 767.3413(3.36E+1)  & --              & 0(0)               & 886.8963(4.88E+0) & ${\dag}$ \\\hline
                       & 3   & \multicolumn{1}{>{\columncolor{mycyan}}c}{\textbf{7.5836(1.83E-2)}}    & 6.8667(1.84E-1)    & 6.8151(1.29E+0) & 6.6662(1.30E-1)    & 4.8432(2.94E-1)   & ${\dag}$ \\\cline{2-8}
\multirow{2}{*}{DTLZ2} & 5   & \multicolumn{1}{>{\columncolor{mycyan}}c}{\textbf{19.5436(5.33E-1)}}   & 16.6564(1.28E+0)   & --              & 15.1439(2.83E+0)   & 10.4404(2.89E+0)  & ${\dag}$ \\\cline{2-8}
                       & 8   & \multicolumn{1}{>{\columncolor{mycyan}}c}{\textbf{152.2602(1.97E+1)}}  & 116.2004(2.08E+1)  & --              & 115.8428(3.09E+1)  & 106.4074(2.47E+0) & ${\dag}$ \\\cline{2-8}
                       & 10  & \multicolumn{1}{>{\columncolor{mycyan}}c}{\textbf{1557.6659(2.04E+1)}} & 1126.4054(8.98E+1) & --              & 1441.8467(1.83E+2) & 420.0209(1.03E+1) & ${\dag}$ \\\hline
                       & 3   & \multicolumn{1}{>{\columncolor{mycyan}}c}{\textbf{7.5632(8.65E-2)}}    & 7.0418(3.11E-1)    & 7.1501(2.32E-1) & 5.9904(4.90E-1)    & 4.3620(2.31E-1)   & ${\dag}$ \\\cline{2-8}
\multirow{2}{*}{DTLZ3} & 5   & \multicolumn{1}{>{\columncolor{mycyan}}c}{\textbf{19.4896(4.89E-1)}}   & 18.2492(1.55E+0)   & --              & 0(0)               & 9.0461(2.67E+0)   & ${\dag}$ \\\cline{2-8}
                       & 8   & \multicolumn{1}{>{\columncolor{mycyan}}c}{\textbf{149.2408(1.38E+1)}}  & 131.1665(1.49E+1)  & --              & 0(0)               & 107.3066(3.96E+0) & ${\dag}$ \\\cline{2-8}
                       & 10  & \multicolumn{1}{>{\columncolor{mycyan}}c}{\textbf{1255.4721(2.58E+1)}} & 1101.5965(7.95E+1) & --              & 0(0)               & 422.5469(1.02E+1) & ${\dag}$ \\\hline
                       & 3   & \multicolumn{1}{>{\columncolor{mycyan}}c}{\textbf{7.5791(1.71E-2)}}    & 7.2202(2.25E-1)    & 6.9368(1.33E+0) & 6.6721(7.33E-2)    & 4.2857(9.57E-1)   & ${\dag}$ \\\cline{2-8}
\multirow{2}{*}{DTLZ4} & 5   & \multicolumn{1}{>{\columncolor{mycyan}}c}{\textbf{19.8557(6.58E-1)}}   & 16.1390(1.24E+0)   & 1.4300(2.26E+0) & 17.3432(7.91E-1)   & 9.4968(2.71E+0)   & ${\dag}$ \\\cline{2-8}
                       & 8   & \multicolumn{1}{>{\columncolor{mycyan}}c}{\textbf{151.3030(3.77E+0)}}  & 115.2255(1.39E+1)  & 0.0724(2.58E-1) & 123.6319(9.93E+0)  & 106.7578(2.42E+0) & ${\dag}$ \\\cline{2-8}
                       & 10  & \multicolumn{1}{>{\columncolor{mycyan}}c}{\textbf{1266.1962(1.87E+1)}} & 1058.6735(2.38E+2) & --              & 1179.5239(1.43E+2) & 442.0123(9.45E+0) & ${\dag}$ \\\hline
\end{tabular}

\begin{tablenotes}
\item[1] $\dag$ denotes the best mean metric value is significantly better than the other peers according to the Wilcoxon's rank sum test at a 0.05 significance level. -- means all solutions obtained by the corresponding algorithm are dominated by the other counterparts, thus no solution can be used for R-metric computation.
\end{tablenotes}
\end{table*}

\begin{table*}[htbp]
\tiny
\centering
\caption{Comparison Results of R-IGD and R-HV Values on the Attainable Aspiration Level Vector}
\label{tab:Rmetric-attainable}

\begin{tabular}{|c|c|c|c|c|c|c|c|}
\cline{3-7}
\multicolumn{1}{c}{}   &     & \multicolumn{5}{c|}{R-IGD} & \multicolumn{1}{c}{} \\\hline
Test Problem           & $m$ & MOEA/D-STM                                                            & R-NSGA-II         & g-NSGA-II         & r-NSGA-II         & NSGA-III          & $s$      \\\hline
                       & 3   & \multicolumn{1}{>{\columncolor{mycyan}}c}{\textbf{9.929E-2(1.09E-2)}} & 1.040E(1.85E-4)   & --                & 2.145E-1(1.54E-1) & 1.175E-1(2.70E-2) & ${\dag}$ \\\cline{2-8}
\multirow{2}{*}{DTLZ1} & 5   & \multicolumn{1}{>{\columncolor{mycyan}}c}{\textbf{2.020E-1(4.41E-4)}} & 3.892E-1(6.57E-3) & 1.106E+3(2.83E+2) & 1.808E+1(1.07E+1) & 3.029E-1(8.56E-2) & ${\dag}$ \\\cline{2-8}
                       & 8   & \multicolumn{1}{>{\columncolor{mycyan}}c}{\textbf{2.933E-1(3.03E-4)}} & 7.669E-1(1.62E-3) & 1.012E+3(2.63E+2) & 7.926E+0(5.30E+0) & 4.072E-1(1.41E-3) & ${\dag}$ \\\cline{2-8}
                       & 10  & \multicolumn{1}{>{\columncolor{mycyan}}c}{\textbf{2.113E-1(1.55E-2)}} & 3.757E-1(1.97E-2) & 2.178E+3(7.01E+2) & 3.001E+1(1.13E+1) & 3.080E-1(2.87E-3) & ${\dag}$ \\\hline
                       & 3   & \multicolumn{1}{>{\columncolor{mycyan}}c}{\textbf{7.282E-2(8.12E-5)}} & 9.257E-2(1.63E-2) & 2.716E+0(1.48E+0) & 4.917E-1(4.02E-1) & 3.322E-1(8.48E-3) & ${\dag}$ \\\cline{2-8}
\multirow{2}{*}{DTLZ2} & 5   & \multicolumn{1}{>{\columncolor{mycyan}}c}{\textbf{2.349E-1(1.23E-3)}} & 9.620E-1(3.32E-2) & 3.106E+0(7.02E-1) & 9.287E-1(8.22E-2) & 7.764E-1(2.89E-1) & ${\dag}$ \\\cline{2-8}
                       & 8   & \multicolumn{1}{>{\columncolor{mycyan}}c}{\textbf{3.975E-1(1.45E-1)}} & 8.642E-1(2.98E-2) & 4.952E+0(1.09E+0) & 8.899E-1(1.98E-1) & 9.650E-1(2.14E-2) & ${\dag}$ \\\cline{2-8}
                       & 10  & \multicolumn{1}{>{\columncolor{mycyan}}c}{\textbf{5.710E-1(6.16E-3)}} & 6.763E-1(1.05E-2) & --                & 6.680E-1(4.56E-2) & 7.201E-1(1.58E-2) & ${\dag}$ \\\hline
                       & 3   & \multicolumn{1}{>{\columncolor{mycyan}}c}{\textbf{7.282E-2(8.12E-5)}} & 9.257E-2(1.63E-2) & 6.229E-2(1.58E-2) & 3.576E-1(2.56E-1) & 3.486E-1(7.75E-3) & ${\dag}$ \\\cline{2-8}
\multirow{2}{*}{DTLZ3} & 5   & \multicolumn{1}{>{\columncolor{mycyan}}c}{\textbf{2.368E-1(1.50E-3)}} & 9.888E-1(4.23E-2) & --                & 2.100E+1(1.29E+1) & 7.513E-1(2.82E-1) & ${\dag}$ \\\cline{2-8}
                       & 8   & \multicolumn{1}{>{\columncolor{mycyan}}c}{\textbf{4.058E-1(1.48E-1)}} & 8.625E-1(3.70E-2) & --                & 5.091E+1(2.48E+1) & 9.680E-1(2.33E-2) & ${\dag}$ \\\cline{2-8}
                       & 10  & \multicolumn{1}{>{\columncolor{mycyan}}c}{\textbf{5.718E-1(5.77E-3)}} & 6.722E-1(1.22E-2) & --                & 4.634E+1(1.55E+1) & 7.154E-1(5.81E-3) & ${\dag}$ \\\hline
                       & 3   & \multicolumn{1}{>{\columncolor{mycyan}}c}{\textbf{1.079E-1(1.03E-1)}} & 8.463E-2(1.34E-2) & 1.137E+1(5.87E+1) & 4.784E-1(3.04E-1) & 3.637E-1(1.38E-1) & ${\dag}$ \\\cline{2-8}
\multirow{2}{*}{DTLZ4} & 5   & \multicolumn{1}{>{\columncolor{mycyan}}c}{\textbf{2.176E-1(2.63E-2)}} & 1.088E+0(1.16E-1) & 2.299E+0(9.99E-1) & 1.125E+0(4.78E-2) & 7.425E-1(2.76E-1) & ${\dag}$ \\\cline{2-8}
                       & 8   & \multicolumn{1}{>{\columncolor{mycyan}}c}{\textbf{3.700E-1(2.04E-2)}} & 8.351E-1(1.63E-2) & 4.731E+0(1.43E+0) & 8.425E-1(2.87E-2) & 9.639E-1(1.98E-2) & ${\dag}$ \\\cline{2-8}
                       & 10  & \multicolumn{1}{>{\columncolor{mycyan}}c}{\textbf{5.786E-1(1.05E-2)}} & 6.903E-1(8.38E-2) & --                & 6.782E-1(5.34E-2) & 7.084E-1(3.80E-3) & ${\dag}$ \\
\hline
\multicolumn{1}{c}{}   &     & \multicolumn{5}{c|}{R-HV} & \multicolumn{1}{c}{} \\\hline
                       & 3   & \multicolumn{1}{>{\columncolor{mycyan}}c}{\textbf{9.5277(2.15E-1)   }} & 9.1690(2.91E-3)   & --               & 9.3537(1.21E+0)   & 10.0742(2.78E-1)  & ${\dag}$ \\\cline{2-8}
\multirow{2}{*}{DTLZ1} & 5   & \multicolumn{1}{>{\columncolor{mycyan}}c}{\textbf{39.2383(6.59E-2)  }} & 29.9628(3.73E-1)  & 0(0)             & 0(0)              & 38.4327(3.57E+0)  & ${\dag}$ \\\cline{2-8}
                       & 8   & \multicolumn{1}{>{\columncolor{mycyan}}c}{\textbf{377.0948(9.72E-1) }} & 168.6195(2.26E+0) & 0(0)             & 13.0027(3.84E+1)  & 307.5456(6.74E+0) & ${\dag}$ \\\cline{2-8}
                       & 10  & \multicolumn{1}{>{\columncolor{mycyan}}c}{\textbf{1206.1236(3.38E+1)}} & 853.7466(3.46E+1) & 0(0)             & 0(0)              & 983.7258(5.53E+1) & ${\dag}$ \\\hline
                       & 3   & \multicolumn{1}{>{\columncolor{mycyan}}c}{\textbf{10.6020(2.63E-3)  }} & 9.7789(3.31E-1)   & 10.4353(1.98E+0) & 6.8350(2.36E+0)   & 7.9283(4.27E-1)   & ${\dag}$ \\\cline{2-8}
\multirow{2}{*}{DTLZ2} & 5   & \multicolumn{1}{>{\columncolor{mycyan}}c}{\textbf{46.5882(1.72E-1)  }} & 17.5653(9.57E-1)  & 0.8553(1.85E+0)  & 18.8126(2.01E+0)  & 21.3570(6.80E+0)  & ${\dag}$ \\\cline{2-8}
                       & 8   & \multicolumn{1}{>{\columncolor{mycyan}}c}{\textbf{473.8128(8.69E+1) }} & 186.8679(1.02E+1) & 0.3761(1.64E+0)  & 191.6834(3.98E+1) & 169.0514(6.45E+0) & ${\dag}$ \\\cline{2-8}
                       & 10  & \multicolumn{1}{>{\columncolor{mycyan}}c}{\textbf{553.1587(7.06E+1) }} & 396.2293(2.54E+1) & --               & 539.9690(7.53E+1) & 411.3842(1.32E+1) & ${\dag}$ \\\hline
                       & 3   & \multicolumn{1}{>{\columncolor{mycyan}}c}{\textbf{10.4353(2.63E-3)  }} & 9.7789(3.31E-1)   & 10.9590(1.91E-1) & 7.5507(1.83E-1)   & 7.2425(3.35E-1)   & ${\dag}$ \\\cline{2-8}
\multirow{2}{*}{DTLZ3} & 5   & \multicolumn{1}{>{\columncolor{mycyan}}c}{\textbf{46.4903(2.02E-1)  }} & 16.9645(1.17E+0)  & --               & 0(0)              & 21.8665(6.62E+0)  & ${\dag}$ \\\cline{2-8}
                       & 8   & \multicolumn{1}{>{\columncolor{mycyan}}c}{\textbf{467.1839(8.72E+1) }} & 190.9587(1.13E+1) & --               & 0(0)              & 172.8418(6.81E+0) & ${\dag}$ \\\cline{2-8}
                       & 10  & \multicolumn{1}{>{\columncolor{mycyan}}c}{\textbf{553.202(6.41E+0)  }} & 399.3820(4.02E+1) & --               & 0(0)              & 417.2917(1.04E+1) & ${\dag}$ \\\hline
                       & 3   & \multicolumn{1}{>{\columncolor{mycyan}}c}{\textbf{10.4854(8.71E-1)  }} & 9.9489(2.49E-1)   & 10.1254(2.11E+0) & 6.7907(2.01E+0)   & 7.6687(1.14E+0)   & ${\dag}$ \\\cline{2-8}
\multirow{2}{*}{DTLZ4} & 5   & \multicolumn{1}{>{\columncolor{mycyan}}c}{\textbf{47.4997(1.61E+0)  }} & 14.6426(2.85E+0)  & 4.3798(4.87E+0)  & 14.0917(1.07E+0)  & 22.0591(6.50E+0)  & ${\dag}$ \\\cline{2-8}
                       & 8   & \multicolumn{1}{>{\columncolor{mycyan}}c}{\textbf{481.3221(1.65E+1) }} & 188.2725(1.12E+1) & 3.1636(9.40E+0)  & 196.4158(9.51E+0) & 169.2001(6.09E+0) & ${\dag}$ \\\cline{2-8}
                       & 10  & \multicolumn{1}{>{\columncolor{mycyan}}c}{\textbf{548.0637(9.96E+0) }} & 379.7303(4.67E+1) & --               & 441.2142(6.43E+1) & 438.4773(9.68E+0) & ${\dag}$ \\\hline
\end{tabular}

\begin{tablenotes}
\item[1] $\dag$ denotes the best mean metric value is significantly better than the other peers according to the Wilcoxon's rank sum test at a 0.05 significance level. -- means all solutions obtained by the corresponding algorithm are dominated by the other counterparts, thus no solution can be used for R-metric computation.
\end{tablenotes}
\end{table*}

%% file: conclusion.tex

\section{Conclusions}
\label{sec:conclusion}

In this paper, we present a systematic method to incorporate the DM's preference information into the decomposition-based EMO methods in either a priori or interactive manner. In particular, the DM's preference information is modeled as the aspiration level vector which represents the DM's expected value on each objective. Our basic idea is a non-uniform mapping scheme that transforms the originally uniformly distributed reference points into a biased distribution. Note that this transformation is implemented by a nonlinear mapping function of which a reference point's position with respect to the pivot point. By these means, the closer to the DM specified aspiration level vector, the more reference points in view of their higher relevance to the DM's preference information. Different from the existing literature, the ROI's size is fully controllable and intuitively understandable according to a quantitative definition. To facilitate the interactive decision making process, our proposed NUMS is not only able to bias the distribution of the reference points towards the ROI, it can also preserve the ones located on the boundary as well, depending on the DM's requirements. By incorporating the NUMS into our recently developed decomposition-based EMO algorithm, i.e., MOEA/D-STM, its effectiveness is validated by proof-of-principle experiments and comparative studies with several state-of-the-art algorithms on a variety of benchmark problems with 2 to 10 objectives. Experimental results fully demonstrate that MOEA/D-STM assisted by the NUMS is able to approximate various ROIs effectively.

As discussed in~\pref{sec:proposal}, our proposed NUMS uses the exponential distribution in its transformation. One direct extension is the application of some other distribution functions to meet the DM's requirements. As discussed in~\pref{sec:relatedworks}, there are several other ways of articulating the DM's preference information. The other extension of this work is the adaptation of the NUMS to other types of preference articulation, e.g., fuzzy linguistic terms. As mentioned in~\cite{BechikhKSG15}, the presence of more than one DM, as known as group decision making, has rarely been studied in the literature. According to the Arrow's impossibility theorem~\cite{Arrow50}, it is challenging to find a consensus decision to which all participating DM agree. In future, it is interesting to generalize the NUMS to accommodate multiple DMs simultaneously. Finally, to further facilitate the interactive process, it is worth considering the combination of human computer interaction techniques~\cite{Qudrat-Ullah06} and the preference-based EMO.